\documentclass[letterpaper]{article} 
\usepackage[preprint]{aaai2027}  
\usepackage[hyphens]{url}  
\usepackage{graphicx} 
\usepackage{natbib}  
\usepackage{caption} 
\usepackage{algorithm}
\usepackage{algorithmic}

\usepackage{colortbl}
\usepackage{tcolorbox}
\tcbuselibrary{skins,breakable}

\usepackage{amsmath} 
\usepackage{amsfonts} 
\usepackage{amsthm}

\usepackage{booktabs} 
\usepackage{multirow} 

\usepackage{xspace}
\newcommand{\model}{\textsc{RelayEvolve}\xspace}

\usepackage{newfloat}
\usepackage{listings}
\DeclareCaptionStyle{ruled}{labelfont=normalfont,labelsep=colon,strut=off} 
\floatstyle{ruled}
\newfloat{listing}{tb}{lst}{}
\floatname{listing}{Listing}

\usepackage{booktabs}

\title{Relay, Don’t Route: \\Adaptive Population Handoff for Cost-Efficient LLM-Driven Evolution}
\author{
    Sichun Luo\textsuperscript{\rm 1}, Yi Huang\textsuperscript{\rm 2}, Guanzhi Deng\textsuperscript{\rm 3}, Haibo Wang\textsuperscript{\rm 4}\\ Haochen Luo\textsuperscript{\rm 1}, Lei Li\textsuperscript{\rm 1}, Zefa Hu\textsuperscript{\rm 2}, Junlan Feng\textsuperscript{\rm 2}, Qi Liu\textsuperscript{\rm 1}
}
\affiliations{
    \textsuperscript{\rm 1}The University of Hong Kong \quad
    \textsuperscript{\rm 2}JIUTIAN Research, China Mobile\\
    \textsuperscript{\rm 3}City University of Hong Kong \quad \textsuperscript{\rm 4}Carnegie Mellon University
\\
\texttt{sichunluo2@gmail.com}
    
}

\begin{document}

\maketitle

\begin{abstract}
Large language model (LLM)-driven evolution has shown promise for program search and algorithm discovery, but relying on strong models throughout long evolutionary runs is costly. A natural alternative is to combine cheap and strong models under a fixed inference budget. However, existing approaches typically allocate models at the level of individual queries or mutation steps, overlooking that evolutionary search is \textit{stateful}: each generated candidate changes the population from which subsequent mutations are produced.

We empirically analyze LLM-driven evolutionary trajectories and find that search progress is strongly front-loaded, early trajectory performance is informative but noisy, and cheap models recover much of the early progress achieved by strong models at lower cost. Motivated by these findings, we propose \textbf{\model}, a training-free framework that shifts budget allocation from individual calls to evolving populations through adaptive \textit{population handoff}. A cheap model explores multiple trajectories in short blocks allocated by a bandit scheduler. Relay Gain, defined as the marginal improvement of a compact, quality-diverse candidate bank constructed for handoff, serves as the scheduler reward and determines when to hand off. The curated candidates initialize a shared strong model population for refinement. Across four benchmarks and three budgets, \model{} achieves the highest mean score in 11 of 12 settings, outperforming competitive baselines. Our results suggest that in stateful search, budget allocation should be organized around the population, not the individual call.
\end{abstract}

\section{Introduction}


Evolutionary search driven by large language models (LLMs) has emerged as a powerful paradigm for algorithm discovery
\cite{romera2024mathematical,liu2024evolution}
and program optimization
\cite{luo2026seaevo,liu2026cognitive}.
These methods iteratively generate, evaluate, and refine candidate programs, with LLMs acting as mutation operators guided by fitness feedback.
Although strong LLMs enable effective mutation, repair, and refinement, relying on them throughout long evolutionary runs is expensive due to the many sequential calls required.
This raises a central question:
\textit{how should a fixed inference budget be allocated over the course of LLM-driven evolution?}

Combining cheap and strong models under a fixed inference budget is a natural approach to reduce cost. 
Prior work has explored cost-aware routing at the request level and model selection within evolutionary search
\cite{ding2024hybrid,ong2024routellm,lange2025shinkaevolve,
ray2026adaptevolve,tanveer2026levi}.
However, these approaches primarily optimize model selection for individual calls or mutation decisions, without explicitly considering how candidates are accumulated and transferred across evolutionary phases.
This local view overlooks a key property of evolutionary search: model calls are state-coupled, because each proposal changes the population inherited by future calls.
Therefore, the key question is  not which model should make the next call, but when a cheap search phase has produced a population worth handing off to a stronger model.

To understand how inference budgets should be allocated, we analyze LLM-driven evolutionary trajectories and identify three patterns. First, fitness improvement is strongly front-loaded, and breakthrough events are concentrated early in the search. Second, early trajectory quality provides useful but noisy signals, suggesting that it alone is insufficient for reliable allocation. Third, cheap models capture a substantial fraction of the early progress achieved by strong models with much lower cost. 
These observations motivate population handoff rather than call-level model selection: a cheap model explores multiple trajectories and constructs a compact, quality-diverse candidate bank, which is then handed to a stronger model for focused refinement.

In this paper, we propose \textbf{\model}, a training-free framework for cost-efficient LLM-driven evolution under a fixed inference budget.
\model first uses a cheap model to explore multiple trajectories in short blocks, with a Grow--Deepen bandit scheduler allocating search effort across trajectories.
The scheduler is guided by Relay Gain, defined as the marginal improvement of an online relay bank that balances solution quality and coverage.
When exploration saturates, \model curates a compact seed set from the candidate pool using greedy submodular selection.
The selected seeds initialize a shared strong model population for focused refinement.
In this way, the relay objective connects trajectory allocation, handoff timing, seed selection, and strong model refinement.
We evaluate \model on four program evolution benchmarks under different budget settings.
Our experiments show that \model improves the cost--performance trade-off over  baseline methods.

Our contributions are summarized as follows:
\begin{itemize}
    \item We empirically characterize LLM-driven evolutionary
    trajectories. We find that progress is strongly front-loaded,
    early trajectory quality provides useful but noisy signals,
    and cheap models capture a substantial fraction of early search
    progress with lower cost.

    \item We introduce \model, a training-free population handoff
    framework that uses Relay Gain to coordinate trajectory allocation
    and handoff timing. A block-level Grow--Deepen bandit scheduler explores
    multiple trajectories, while greedy submodular curation selects a
    compact and complementary seed population for strong model
    refinement.

    \item Across four program evolution benchmarks and three budget
levels, \model demonstrates the effectiveness of population
handoff, producing improved cost--performance trajectories and
achieving the highest mean score in 11 of 12 benchmark--budget
settings.
\end{itemize}

\section{Related Work}

\paragraph{LLM-driven evolutionary search.}
Large language models have been used as semantic variation operators
for prompt, heuristic, and program search
\cite{guo2024connecting,ye2024reevo,agrawal2025gepa,yan2026pacevolve,qu2026coral,luo2026harness}.
FunSearch paired an LLM with automated evaluation and population-based
selection to discover mathematical constructions and algorithms
\cite{romera2024mathematical}, while AlphaEvolve extended this paradigm
to larger codebases and a broader range of scientific and systems
problems \cite{novikov2025alphaevolve}. Subsequent frameworks improve
the evolutionary harness through mechanisms such as adaptive parent
sampling, novelty rejection, archive management, and reflective
mutation \cite{lange2025shinkaevolve,assumpccao2025codeevolve}.
These works establish increasingly capable search operators and
population-management mechanisms. Our focus is instead on how a fixed
inference budget should be allocated between cheap and strong
models over the course of a stateful evolutionary search.

\paragraph{Cost-aware LLM routing and evolutionary search allocation.}
Cost-aware LLM systems allocate inference across models with different
capabilities and costs. FrugalGPT constructs request-level model
cascades \cite{chen2023frugalgpt}, while RouteLLM learns to route
individual queries between weaker and stronger models
\cite{ong2024routellm}. Multi-model allocation has also been explored
within evolutionary search: ShinkaEvolve uses a bandit-based LLM
ensemble selector \cite{lange2025shinkaevolve}, AdaptEvolve escalates
individual refinement steps according to generation confidence
\cite{ray2026adaptevolve}, and LEVI routes different mutation roles to
different model classes \cite{tanveer2026levi}. A related line of work
allocates computation across search processes rather than models.
Hyperband uses successive halving to distribute resources among
configurations \cite{li2018hyperband}, while AdaEvolve adaptively
allocates computation across evolving populations based on search
progress \cite{cemri2026adaevolve}. These approaches primarily optimize
local decisions: selecting a model for an individual step or allocating
additional computation to a search process. In contrast, \model
optimizes the population transferred between model phases: a cheap model
constructs a compact candidate bank, and its marginal set-level
improvement jointly guides trajectory allocation, stopping, and the
eventual handoff to a shared strong model population.

\section{Empirical Motivation}
\label{sec:empirical-motivation}

We examine how LLM-driven evolutionary search progresses over a fixed
horizon of $T=100$ generations. We consider three tasks: Circle Packing (Square),
Transaction Scheduling, and Prism~\cite{liu2026skydiscover}. For each
task, we independently run a cheap model
(\texttt{Qwen-3.5-Flash}~\cite{openrouter_qwen35_flash}) and a strong
model (\texttt{Qwen-3.5-Plus}~\cite{openrouter_qwen35_plus}) using the
same ShinkaEvolve backend~\cite{lange2025shinkaevolve}. We conduct ten
independent runs for each combination of model and task, resulting in
60 runs in total. 

\begin{figure}[!t]
    \centering
\includegraphics[width=0.5\textwidth]{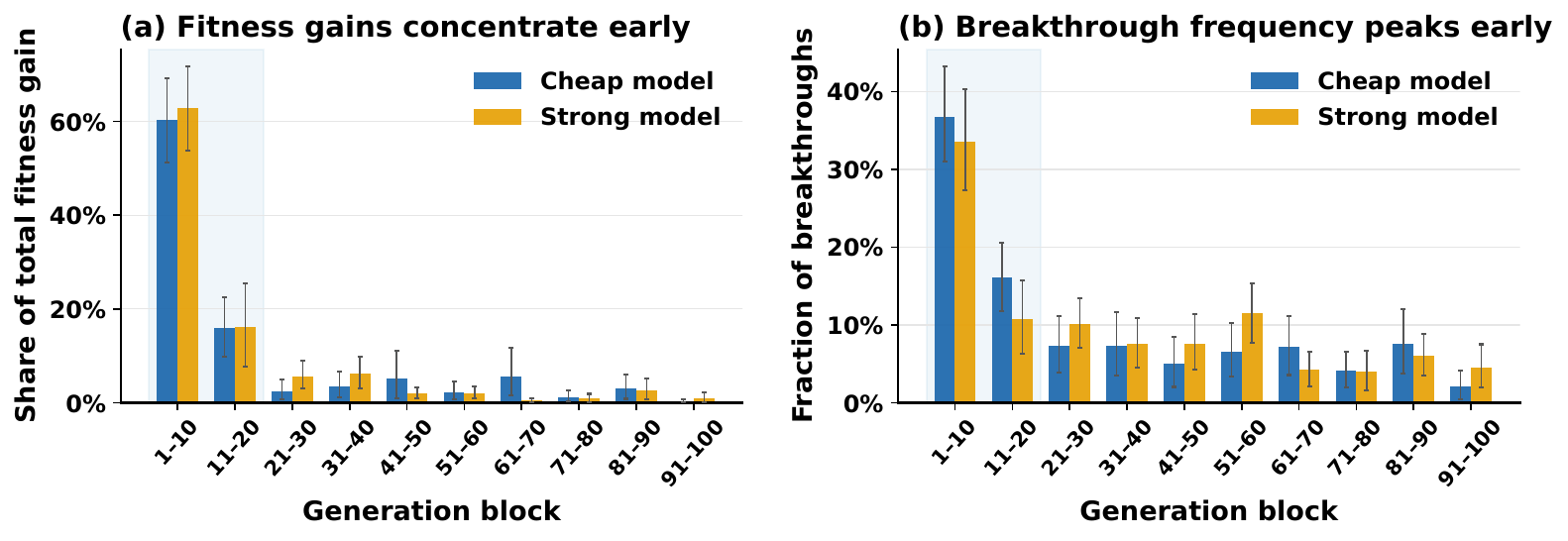}
    \caption{
(a) Fraction of each run's eventual best-so-far fitness improvement
contributed by each ten-generation bin.
(b) Fraction of breakthrough events occurring in each bin.
Bars show means over three tasks and ten independent runs per model.
Error bars denote 95\% bootstrap confidence intervals.
The shaded region marks the first 20 generations.
    }
\label{fig:f1}

\end{figure}

\begin{figure}[!t]
    \centering
\includegraphics[width=0.48\textwidth]{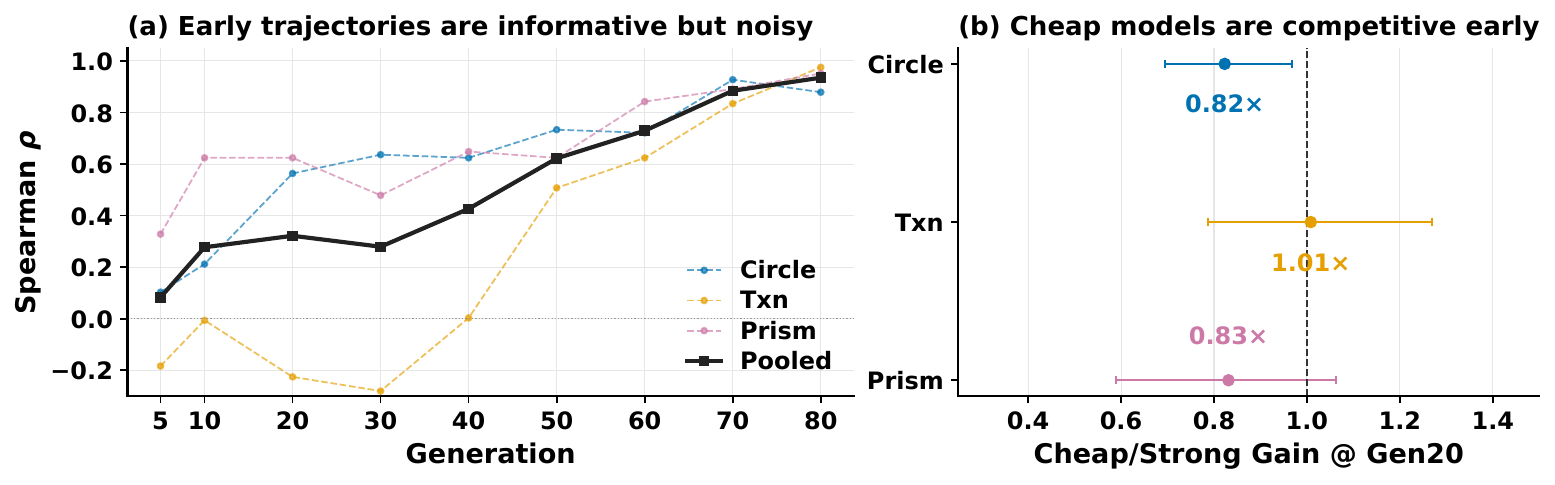}
    \caption{
(a) Spearman correlation between cheap model trajectory ranks at
generation $K$ and their final ranks at generation 100.
Early signals are informative but task-dependent and noisy.
(b) Ratio of cheap to strong model improvement over the first 20
generations. Error bars denote 95\% bootstrap confidence intervals.
    }
\label{fig:f2}

\end{figure}

\paragraph{Finding 1: Search progress is strongly front-loaded.}
For each run, we divide the trajectory into ten-generation bins and
measure the fraction of total best-so-far improvement contributed by
each bin. We also measure the temporal distribution of breakthrough
events. As shown in Figure~\ref{fig:f1}(a), the first 20 generations
account for $76.3\%$ and $79.1\%$ of the total improvement achieved by
the cheap and strong models, respectively. Figure~\ref{fig:f1}(b) shows
that breakthrough events are also concentrated early in the search.
Moreover, each run contains a median of only seven breakthroughs,
indicating that progress typically occurs through a small number of
discrete jumps rather than through steady incremental gains. 

\paragraph{Finding 2: Early cheap model trajectories are partially
predictive but noisy.}
Within each task, we rank the ten cheap model trajectories by their
best-so-far fitness at generation $K$ and compute the Spearman
correlation with their rankings at generation $T$. As shown in
Figure~\ref{fig:f2}(a), early trajectory quality provides a weak  signal of final performance. The pooled correlation is
$\rho=0.28$ at $K=10$ and $\rho=0.32$ at $K=20$, increasing to
$\rho=0.62$ at $K=50$. However, the task-level correlations vary
substantially and can even be negative at early horizons. Early fitness
should therefore be treated as a soft allocation signal rather than as
a criterion for committing to a single trajectory.

\paragraph{Finding 3: The cheap model captures substantial early
progress.}
We compare the best-so-far improvement achieved by the cheap and strong
models during the first 20 generations. As shown in
Figure~\ref{fig:f2}(b), the ratio of cheap model to strong model
improvement is $0.82\times$, $1.01\times$, and $0.83\times$ on Circle
Packing, Transaction Scheduling, and Prism, respectively. Although the
cheap model does not uniformly match the strong model, it captures a
substantial fraction of early progress at lower inference cost. 
This
suggests that strong model calls need not be used uniformly from the
beginning of the search.

\begin{figure*}[!t]
    \centering
\includegraphics[width=0.95\textwidth]{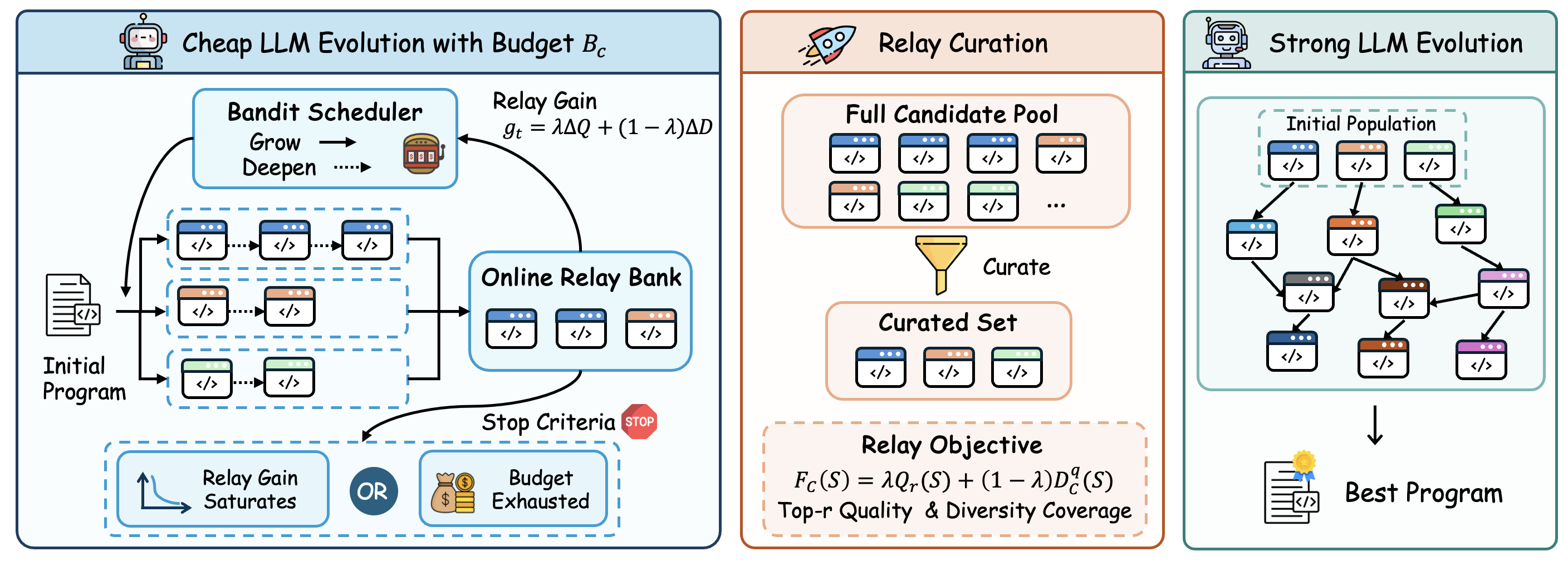}
    \caption{
Overview of \model. A cheap LLM explores multiple trajectories in short blocks allocated by a
Grow--Deepen bandit, whose reward is the Relay Gain of an online
quality-diverse bank. Cheap model evolution stops when Relay Gain saturates or its
budget is exhausted. \model then reruns greedy submodular curation on the
full candidate pool and hands the resulting compact seed set to a shared
strong LLM population for focused refinement with the remaining budget.
    }
\label{fig:pipeline}
\end{figure*}

\paragraph{Implications for \model.}
These findings motivate a population handoff strategy.
The rapid decline in improvement motivates short-block scheduling and
adaptive stopping. The instability of early rankings motivates
maintaining multiple trajectories rather than committing to one early
leader. The substantial early progress achieved by the cheap model
motivates reserving strong model budget for later refinement. \model
therefore evaluates cheap model actions through the marginal
improvement of a compact relay bank that balances quality and coverage.
This signal determines whether to start a new trajectory or extend an
existing one, as well as when to terminate cheap model search. The same
objective form is then used to select a complementary seed population
for subsequent strong model evolution.



\section{Problem Formulation}

We consider an LLM-driven evolutionary search that maximizes a task-specific
fitness function $f$. At each generation, an LLM mutates a previously
evaluated program, and an external evaluator scores the resulting candidate.
Let $\mathcal{C}_t$ denote the set of valid programs discovered after $t$ model
calls, and let
\[
f_t^* = \max_{x \in \mathcal{C}_t} f(x)
\]
denote the best fitness obtained up to that point.

We have access to a cheap model $m_c$ and a strong model $m_s$. Their realized
inference costs depend on token usage, but calls to $m_c$ are substantially
cheaper in expectation. In addition to the monetary inference budget, we impose
a maximum number of model calls to account for the computational cost of
candidate evaluation. Given a total inference
budget $B$ and a call budget $N$, our objective is
\begin{equation}
\max_{\pi}\ \mathbb{E}[f_T^*]
\quad
\text{s.t.}
\quad
\sum_{t=1}^{T} c_t \le B,
\qquad
T \le N,
\label{eq:objective}
\end{equation}
where $c_t$ is the realized inference cost of the model called at step $t$, 
and $\pi$ denotes a search policy that governs model usage and resource
allocation throughout the evolutionary process.

\section{Method}
\label{sec:method}

\subsection{Overview of \model}
\label{sec:method-overview}

Figure~\ref{fig:pipeline} illustrates \model, a training-free framework
that allocates a fixed inference budget through adaptive population handoff.
Instead of selecting a model independently for each mutation, \model first
uses a cheap model to explore multiple trajectories in short, fixed-length
blocks. After each block, it updates an online \emph{relay bank} containing
a compact set of high-quality and complementary candidates.

We define \emph{Relay Gain} as the marginal improvement of the relay bank.
A Grow--Deepen bandit scheduler uses this block-level reward to decide
whether to start a new trajectory or extend an existing one. The handoff
from cheap model exploration to strong model refinement is triggered either
by persistent low Relay Gain or by exhaustion of the cheap model budget.
At handoff, \model re-optimizes the same relay objective over the complete
cheap model candidate pool to construct a compact seed population. The
selected seeds initialize a single shared strong model population, which
uses the remaining budget for refinement. Thus, the same population-level
objective coordinates three decisions: how to allocate cheap model search,
when to terminate cheap model exploration, and which candidates to hand off
to the strong model.

\subsection{Relay Objective and Relay Gain}
\label{sec:relay-objective}

Cheap model search proceeds in fixed-length blocks, with each block
advancing one trajectory for $h$ generations. Before the $t$-th block,
\model maintains a set of active trajectories $\mathcal{I}_t$, a
deduplicated candidate pool $\mathcal{C}_t$ containing all candidates
discovered so far, and an online relay bank
$S_t\subseteq\mathcal{C}_t$ with $|S_t|\leq k$.
The relay bank is a compact population intended for downstream handoff.

To define how a relay bank is evaluated, consider any fixed candidate
pool $\mathcal{C}$ and a bank $S\subseteq\mathcal{C}$.
Motivated by the quality-diversity principle of maintaining
high-performing yet diverse solution sets
\citep{mouret2015illuminating,pugh2016quality}, we score $S$
according to two complementary properties: the top-$r$ quality of its selected
candidates and its diversity coverage of promising regions in $\mathcal{C}$.

\paragraph{Top-$r$ quality.}
Let $q(x)\in[0,1]$ denote the task-specific normalized fitness of
candidate $x$, with higher values indicating better solutions.
Let $q_{(j)}(S)$ be the $j$-th largest normalized quality in $S$, with
$q_{(j)}(S)=0$ for $j>|S|$. We define
\begin{equation}
\label{eq:top-r-quality}
Q_r(S)
=
\frac{1}{r}
\sum_{j=1}^{r} q_{(j)}(S).
\end{equation}
This term rewards banks containing multiple high-quality anchors without
requiring every bank slot to contain one of the highest-scoring candidates.

\paragraph{Diversity coverage.}
Each program candidate $x$ is represented using a code embedding
$e_{\mathrm{code}}(x)$ and a textual metadata embedding
$e_{\mathrm{text}}(x)$. We combine the two views through
\begin{equation}
\label{eq:candidate-similarity}
\begin{split}
\operatorname{sim}(x,x')
&=
\operatorname{clip}\Bigl(
\eta\,
\cos\!\left(
e_{\mathrm{code}}(x),
e_{\mathrm{code}}(x')
\right)
\\
&+
(1-\eta)\,
\cos\!\left(
e_{\mathrm{text}}(x),
e_{\mathrm{text}}(x')
\right),
0,1
\Bigr),
\end{split}
\end{equation}
where $\eta\in[0,1]$ controls the contribution of the code view.

For a fixed candidate pool $\mathcal{C}$, we define the quality-weighted diversity
coverage of bank $S$ as
\begin{equation}
\label{eq:quality-coverage}
D_{\mathcal{C}}^q(S)
=
\frac{
\sum_{v\in\mathcal{C}}
q(v)
\max_{x\in S}
\operatorname{sim}(v,x)
}{
\sum_{v\in\mathcal{C}}q(v)
}.
\end{equation}
We set $D_{\mathcal{C}}^q(S)=0$ when
$\sum_{v\in\mathcal{C}}q(v)=0$, and define the maximum over an empty bank
as zero. This facility-location term favors banks that cover the
high-quality mass of the candidate pool. Weighting candidates by $q(v)$
prevents limited bank capacity from being spent primarily on diverse but
low-quality outliers.

\paragraph{Relay objective.}
The relay value of $S$ with respect to reference pool $\mathcal{C}$ is
\begin{equation}
\label{eq:relay-objective}
F_{\mathcal{C}}(S)
=
\lambda Q_r(S)
+
(1-\lambda)D_{\mathcal{C}}^q(S),
\qquad
\lambda\in[0,1].
\end{equation}
The first term retains high-quality anchors, while the second encourages
coverage of complementary high-quality regions. For a fixed reference pool,
both terms are monotone submodular, and therefore
$F_{\mathcal{C}}$ is also monotone submodular. 

\paragraph{Online bank update.}
Suppose block $t$ produces a set $X_t$ of valid candidates. We first update
the deduplicated reference pool:
\begin{equation}
\label{eq:candidate-pool-update}
\mathcal{C}_{t+1}
=
\operatorname{Dedup}
\left(
\mathcal{C}_t\cup X_t
\right),
\end{equation}
where $\operatorname{Dedup}$ removes duplicate programs from the pool.

Starting from $S_t$, we process candidates in $X_t$ in generation order.
A candidate is added directly if the bank is not full. Otherwise, it replaces
the bank element whose removal yields the largest increase in
$F_{\mathcal{C}_{t+1}}$, provided that the replacement strictly improves the
objective. The resulting bank is denoted by $S_{t+1}$. 

We define the \emph{Relay Gain} of block $t$ as
\begin{equation}
\label{eq:relay-gain}
g_t
=
F_{\mathcal{C}_{t+1}}(S_{t+1})
-
F_{\mathcal{C}_{t+1}}(S_t)
\geq 0.
\end{equation}
Both banks are evaluated against the same updated reference pool
$\mathcal{C}_{t+1}$. This ensures that $g_t$ measures improvement of the
population available for handoff rather than a change in the underlying
reference pool.

Relay Gain is therefore a block-level and set-level reward. Unlike immediate
best-fitness improvement, it credits a block for adding either higher-quality
candidates or candidates that complement the current handoff population.

\subsection{Adaptive Cheap Model Search}
\label{sec:cheap-search}

\paragraph{Grow--Deepen actions.}
Cheap model search proceeds in blocks of $h$ generations. Each block applies
one of two actions:
\begin{itemize}
    \item \textsc{Grow} starts a new trajectory from the common initial
    program and executes its first block.
    \item \textsc{Deepen}$(i)$ resumes an existing trajectory $i$ for one
    additional block.
\end{itemize}
Independent model sampling creates variation across \textsc{Grow}
trajectories despite their shared initialization. We limit both the total
number of active trajectories and the horizon of each trajectory, preventing
a single search region from consuming the entire cheap model budget.

Because the significance of a fixed absolute gain depends on the current
relay bank value, we normalize $g_t$ into a bounded relative Relay Gain:
\begin{equation}
\label{eq:relative-relay-gain}
\rho_t
=
\operatorname{clip}
\left(
\frac{
g_t
}{
\max
\left\{
F_{\mathcal{C}_{t+1}}(S_t),
\epsilon_F
\right\}
},
0,
1
\right),
\end{equation}
where $\epsilon_F>0$ is a stabilizing floor. We use $g_t$ to report absolute
bank improvement and $\rho_t$ as the reward for scheduling and adaptive
handoff.

\paragraph{Grow--Deepen bandit scheduler.}
We treat \textsc{Grow} as a shared meta-arm and associate one
\textsc{Deepen} arm with each active trajectory. Search begins with a small
number of independent \textsc{Grow} blocks that initialize the trajectory
set and relay bank. Their rewards are excluded from the scheduler history
because gains observed while filling an initially empty bank are strongly
affected by candidate arrival order.

After initialization, a \textsc{Grow} reward updates the shared
\textsc{Grow} statistics and provides the initial utility estimate for the
new trajectory. A \textsc{Deepen}$(i)$ reward updates only the statistics of
trajectory $i$.

Because the utility of a trajectory can change as search progresses, we use a
recent-window UCB-style score \cite{besbes2014stochastic}. Let $n_a(t)$ be the number of reward
observations associated with action $a$, and let
$\widehat{\mu}^{\,\mathrm{recent}}_a(t)$ be the mean of its most recent $w$
rewards. We define
\begin{equation}
\label{eq:recent-ucb}
U_a(t)
=
\begin{cases}
+\infty,
&
n_a(t)=0,
\\[2mm]
\widehat{\mu}^{\,\mathrm{recent}}_a(t)
+
c
\sqrt{
\displaystyle
\frac{
\log\!\left(\max\{2,t+1\}\right)
}{
n_a(t)
}
},
&
n_a(t)>0,
\end{cases}
\end{equation}
where $c$ controls exploration. At each block, \model selects the available
action with the largest score. \textsc{Grow} becomes unavailable after the
trajectory limit is reached, and \textsc{Deepen}$(i)$ becomes unavailable
once trajectory $i$ reaches its horizon.

All actions use the same cheap model and block length, so the scheduler does
not further normalize rewards by nominal action cost. Realized generation
and embedding costs are nevertheless charged to the global inference budget.

\paragraph{Adaptive handoff.}
Relative Relay Gain also determines when cheap model exploration terminates.
Let $\epsilon_{\mathrm{rel}}$ be a saturation threshold and $p$ a patience
window. We trigger handoff when
\begin{equation}
\label{eq:relay-saturation}
\max_{j=t-p+1,\ldots,t}
\rho_j
<
\epsilon_{\mathrm{rel}}.
\end{equation}
Thus, handoff is triggered only when every block in the recent window contributes less than the required relative improvement. The cheap phase also terminates if its stage budget $B_c$ is exhausted.

\subsection{Population Handoff and Strong Model Refinement}
\label{sec:handoff}

The online relay bank provides efficient streaming rewards, but its contents
can depend on candidate arrival order. Let $\tau$ denote the block at which
cheap model exploration terminates. At handoff, \model therefore re-optimizes
the relay objective over the complete and fixed terminal candidate pool
$\mathcal{C}_{\tau}$.

We first obtain an offline seed set by greedy maximization followed by
objective-improving single-element swaps:
\begin{equation}
\label{eq:offline-curation}
S_{\mathrm{g}}
=
\operatorname{LocalSearch}
\left(
\operatorname{GreedySelect}
\left(
\mathcal{C}_{\tau},
F_{\mathcal{C}_{\tau}},
k
\right)
\right).
\end{equation}
We also apply the same local search to the terminal online bank:
\begin{equation}
\label{eq:online-curation}
S_{\mathrm{o}}
=
\operatorname{LocalSearch}(S_{\tau}).
\end{equation}
The final handoff population is
\begin{equation}
\label{eq:final-curation}
S^*
=
\arg\max_{S\in\{S_{\mathrm{g}},S_{\mathrm{o}}\}}
F_{\mathcal{C}_{\tau}}(S).
\end{equation}

The offline initialization removes arrival-order dependence, while the online
initialization preserves candidates accumulated under the streaming policy.
Because $F_{\mathcal{C}_{\tau}}$ is monotone submodular, greedy selection
under the cardinality constraint $|S|\leq k$ achieves the standard
$(1-1/e)$ approximation guarantee
\cite{nemhauser1978}. Subsequent objective-improving swaps and the final
comparison do not decrease the objective, so $S^*$ preserves this guarantee.

After handoff, all candidates in $S^*$ initialize a single shared
strong model evolutionary population. They are not assigned separate
inference budgets or evolved as independent searches. Instead, the
strong model evolutionary process can select seeds or their descendants
as parents, allowing discoveries originating from different cheap-model
trajectories to interact within a common population. The strong model
uses the remaining inference budget, and \model returns the best valid
program found across both cheap and strong phases.

\section{Experiments}




\newcommand{\avgstd}[2]{#1$_{\scriptsize \pm \text{#2}}$}
\newcommand{\bavgstd}[2]{\textbf{#1}$_{\scriptsize \pm \text{#2}}$}

\definecolor{relay}{gray}{0.85}

\begin{table*}[!t]
\centering
\resizebox{\textwidth}{!}{
\begin{tabular}{lcccccccc}
\toprule
\multirow{2}{*}{\textbf{Method}}
& \multicolumn{2}{c}{\textbf{Circle Packing (Square)}}
& \multicolumn{2}{c}{\textbf{Circle Packing (Rect)}}
& \multicolumn{2}{c}{\textbf{TXN Scheduling}}
& \multicolumn{2}{c}{\textbf{Prism}} \\
\cmidrule(lr){2-3}
\cmidrule(lr){4-5}
\cmidrule(lr){6-7}
\cmidrule(lr){8-9}
& Avg & Best & Avg & Best & Avg & Best & Avg & Best \\
\midrule

\multicolumn{9}{l}{\textit{Budget: 50\%}} \\
All-cheap
& \avgstd{2.3214}{0.1660} & 2.5059
& \avgstd{2.2649}{0.0936} & 2.3471
& \avgstd{3873.89}{129.43} & 3984.06
& \bavgstd{26.2440}{0.0110} & \textbf{26.2560} \\
All-strong
& \avgstd{2.4084}{0.0237} & 2.4347
& \avgstd{2.3471}{0.0120} & 2.3580
& \avgstd{3606.28}{52.11} & 3636.36
& \avgstd{26.0920}{0.2490} & \textbf{26.2560} \\
Fixed-switch
& \avgstd{2.3346}{0.2266} & 2.4769
& \avgstd{2.2681}{0.0674} & 2.3425
& \avgstd{3703.94}{36.47} & 3745.32
& \avgstd{26.0110}{0.2560} & 26.2010 \\
Random
& \avgstd{2.3531}{0.1001} & 2.4637
& \avgstd{2.3207}{0.0588} & 2.3564
& \avgstd{3765.18}{77.17} & 3816.79
& \avgstd{26.1620}{0.0680} & 26.2390 \\
Bandit
& \avgstd{2.2661}{0.0158} & 2.2819
& \avgstd{2.3293}{0.0195} & 2.3501
& \avgstd{3584.00}{147.46} & 3731.34
& \avgstd{26.0580}{0.2690} & 26.2410 \\
LEVI
& \avgstd{2.3627}{0.1238} & 2.4598
& \avgstd{2.2573}{0.0380} & 2.2977
& \avgstd{3697.17}{196.03} & 3831.42
& \avgstd{25.5628}{0.5708} & 26.2218 \\
\rowcolor{relay}\model
& \bavgstd{2.4136}{0.1228} & \textbf{2.5364}
& \bavgstd{2.3599}{0.0069} & \textbf{2.3648}
& \bavgstd{3875.09}{153.52} & \textbf{4000.00}
& \avgstd{26.2408}{0.0131} & \textbf{26.2560} \\
\midrule

\multicolumn{9}{l}{\textit{Budget: 75\%}} \\
All-cheap
& \avgstd{2.3214}{0.1660} & 2.5059
& \avgstd{2.2649}{0.0936} & 2.3471
& \avgstd{3873.89}{129.43} & 3984.06
& \avgstd{26.2440}{0.0110} & \textbf{26.2560} \\
All-strong
& \avgstd{2.4400}{0.0184} & 2.4611
& \avgstd{2.3471}{0.0120} & 2.3580
& \avgstd{3716.89}{147.91} & 3802.28
& \avgstd{26.0930}{0.2490} & \textbf{26.2560} \\
Fixed-switch
& \avgstd{2.3662}{0.1897} & 2.4980
& \avgstd{2.2814}{0.0597} & 2.3502
& \avgstd{3708.46}{31.92} & 3745.32
& \avgstd{26.1910}{0.1130} & \textbf{26.2560} \\
Random
& \avgstd{2.4295}{0.0764} & 2.5006
& \avgstd{2.3227}{0.0606} & 2.3590
& \avgstd{3783.72}{59.20} & 3831.42
& \avgstd{26.1950}{0.0780} & \textbf{26.2560} \\
Bandit
& \avgstd{2.3504}{0.0422} & 2.3952
& \avgstd{2.3382}{0.0233} & 2.3540
& \avgstd{3606.01}{157.82} & 3773.58
& \avgstd{26.1840}{0.0910} & 26.2410 \\
LEVI
& \avgstd{2.4144}{0.0415} & 2.4598
& \avgstd{2.2573}{0.0380} & 2.2977
& \avgstd{3697.17}{196.03} & 3831.42
& \avgstd{25.5628}{0.5708} & 26.2218 \\
\rowcolor{relay}\model
& \bavgstd{2.5246}{0.0408} & \textbf{2.5581}
& \bavgstd{2.3616}{0.0040} & \textbf{2.3648}
& \bavgstd{3875.09}{153.52} & \textbf{4000.00}
& \bavgstd{26.2484}{0.0131} & \textbf{26.2560} \\
\midrule

\multicolumn{9}{l}{\textit{Budget: 100\%}} \\
All-cheap
& \avgstd{2.3214}{0.1660} & 2.5059
& \avgstd{2.2649}{0.0936} & 2.3471
& \avgstd{3873.89}{129.43} & 3984.06
& \avgstd{26.2440}{0.0110} & \textbf{26.2560} \\
All-strong
& \avgstd{2.4453}{0.0290} & 2.4769
& \avgstd{2.3471}{0.0120} & 2.3580
& \avgstd{3721.72}{152.27} & 3816.79
& \avgstd{26.1040}{0.2580} & \textbf{26.2560} \\
Fixed-switch
& \avgstd{2.3926}{0.2108} & 2.5349
& \avgstd{2.2814}{0.0597} & 2.3502
& \avgstd{3780.98}{123.49} & 3921.57
& \avgstd{26.1910}{0.1130} & \textbf{26.2560} \\
Random
& \avgstd{2.4558}{0.0400} & 2.5006
& \avgstd{2.3565}{0.0030} & 2.3590
& \avgstd{3818.38}{95.06} & 3906.25
& \avgstd{26.2020}{0.0820} & \textbf{26.2560} \\
Bandit
& \avgstd{2.4084}{0.0817} & 2.5006
& \avgstd{2.3451}{0.0114} & 2.3540
& \avgstd{3663.03}{159.10} & 3846.15
& \avgstd{26.1940}{0.0810} & 26.2480 \\
LEVI
& \avgstd{2.4144}{0.0415} & 2.4598
& \avgstd{2.2573}{0.0380} & 2.2977
& \avgstd{3697.17}{196.03} & 3831.42
& \avgstd{25.5628}{0.5708} & 26.2218 \\
\rowcolor{relay}\model
& \bavgstd{2.5364}{0.0218} & \textbf{2.5581}
& \bavgstd{2.3616}{0.0040} & \textbf{2.3648}
& \bavgstd{3942.61}{50.29} & \textbf{4000.00}
& \bavgstd{26.2484}{0.0131} & \textbf{26.2560} \\
\bottomrule
\end{tabular}
}
\caption{Performance across four program evolution benchmarks.
All methods report mean score and standard deviation over three runs.
Best denotes the best score achieved.
Best results in each budget--benchmark column are bolded.
}
\label{tab:main_results}
\end{table*}


\subsection{Experimental Setup}

Similar to prior work~\cite{cemri2026adaevolve,liu2026evox}, we evaluate on four algorithm discovery and system optimization tasks: Circle Packing (Square), Circle Packing (Rectangle), TXN Scheduling, and Prism~\cite{liu2026skydiscover}.
We use \texttt{Qwen-3.5-Flash} as the cheap model $m_c$ and \texttt{Qwen-3.5-Plus} as the strong model $m_s$.
The input/output prices are \$0.065/\$0.26 per million tokens for \texttt{Qwen-3.5-Flash} and \$0.26/\$1.56 per million tokens for \texttt{Qwen-3.5-Plus}. Every model call is charged according to its actual input and output token usage.
All methods use the same ShinkaEvolve backend~\cite{lange2025shinkaevolve}, evaluator, mutation template, population size, and context budget. We report the mean and standard deviation over three independent runs.







All methods operate under the same total budget
$B = \beta\, B_{\text{all-strong}}$, where $B_{\text{all-strong}}$ is
the measured cost of a standard 100 generation all-strong run on each
task and $\beta \in \{0.5, 0.75, 1.0\}$. 
To control the computational overhead of evolutionary evaluation, we
cap each run at $N=200$ generations, following the generation budget
constraint in Eq.~(\ref{eq:objective}).
A run
terminates when it exhausts either its budget or the generation cap.
In particular, a full 200 generation cheap run costs less than
$0.5\,B_{\text{all-strong}}$, so All-cheap results coincide across
budget levels. 

\subsection{Baseline Methods}

We compare \model against several representative baselines. 
\textbf{All-cheap} and \textbf{All-strong} use the cheap and strong models,
respectively, throughout the entire search. 
\textbf{Fixed-switch} begins with a fixed cheap model exploration phase and
then switches to the strong model for the remainder of the search
\cite{bhan2026new}. 
\textbf{Random} independently selects between the cheap and strong models at
each generation. 
\textbf{Bandit} treats the two models as arms of a multi-armed bandit and
uses the realized improvement in best-so-far fitness as the reward for model
selection \cite{lange2025shinkaevolve}. 
\textbf{LEVI} assigns different mutation roles to different model classes
through a predefined role-based routing strategy \cite{tanveer2026levi}.
We exclude \textbf{AdaptEvolve} \cite{ray2026adaptevolve} because its
model-selection mechanism requires access to token-level logits, which are
unavailable through the online API services used in our experiments.
All evaluated methods share the same evolutionary backend and are subject to
the same inference cost and model call constraints.

\subsection{Main Results}

Table~\ref{tab:main_results} summarizes performance across four tasks
and three budget levels. \model achieves the highest mean score in 11
of the 12 benchmark--budget settings and attains or ties the best
single-run score on every benchmark. The only non-best mean occurs on
Prism at the lowest budget, where most methods are already near
saturation and the gap is negligible.

The single-model baselines exhibit complementary strengths:
All-strong performs better on Circle Packing, whereas All-cheap is more
competitive on TXN Scheduling and Prism. This task dependence makes a
single predetermined allocation unreliable. By adapting the handoff to
the evolving candidate population, \model matches or outperforms the
stronger single-model baseline in most settings. It also
outperforms Fixed-switch throughout, indicating that the appropriate
transition point depends on search progress rather than a fixed
schedule. Bandit routing is less consistent, suggesting that immediate
call-level fitness improvements can be a myopic signal of a candidate's
downstream value in stateful evolutionary search.

\begin{figure}[!t]
    \centering
\includegraphics[width=0.48\textwidth]{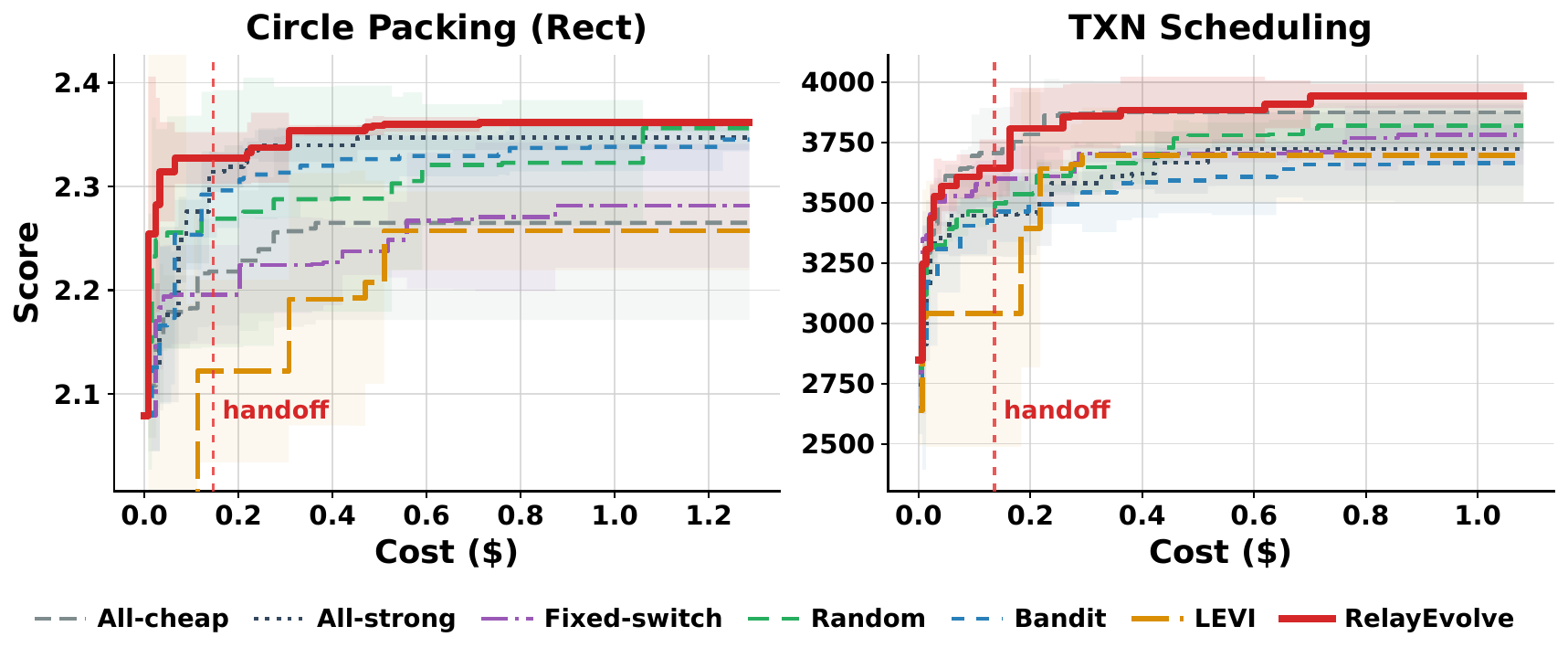}
    \caption{
Cost–performance comparison on Circle Packing (Rect) and TXN Scheduling. Results are averaged over three independent runs, with shaded regions denoting ±1 standard deviation. The vertical dashed line indicates \model's model handoff point.
    }
\label{fig:curve}
\end{figure}

Figure~\ref{fig:curve} illustrates this behavior on two representative
tasks. \model initially follows the rapid progress of All-cheap, then
continues improving after handing the curated population to the strong
model. The resulting trajectory supports the central design of
\model: cheap models are effective for broad early exploration, while
strong models provide greater value when refining a compact population
of promising and complementary candidates.





\begin{figure*}[!t]
    \centering
\includegraphics[width=0.98\textwidth]{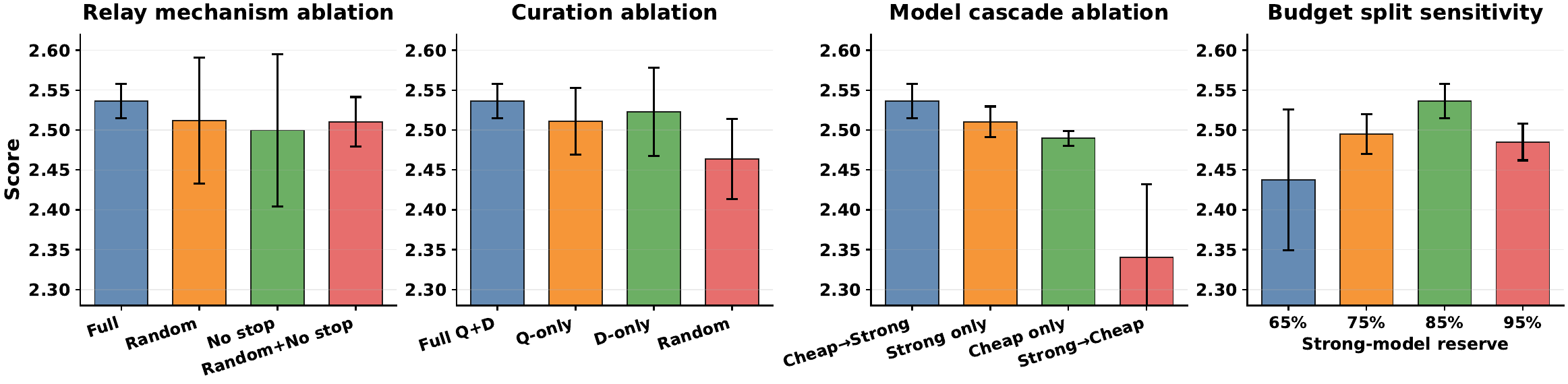}
    \caption{
Ablation studies on Circle Packing (Square) at the 100\% budget
level. From left to right: \textbf{(a)} relay mechanism (relay-gain
allocation and stopping vs.\ random allocation and no stopping);
\textbf{(b)} curation objective (full $Q + D$ vs.\ quality-only,
diversity-only, and random seeds); \textbf{(c)} model cascade direction;
\textbf{(d)} sensitivity to the strong model budget reserve. Bars show
means over three runs, and error bars denote standard deviations.
    }
\label{fig:ablation}
\end{figure*}

\subsection{Ablation Study}
\label{sec:ablation}

As shown in Figure~\ref{fig:ablation},
we conduct ablations on Circle Packing (Square), where the optimization
landscape is sufficiency challenging to reveal the effect of different
design choices. 

\paragraph{Effect of the relay mechanism.}
We first isolate the contribution of the relay-gain mechanism that jointly
controls allocation and stopping decisions. Removing relay-gain based
control consistently reduces performance, while replacing Grow/Deepen
decisions with random choices or removing relay-based stopping leads to
further degradation. These results support our design choice of using a
shared marginal reward signal to coordinate online computation allocation
and termination.

\paragraph{Effect of relay curation.}
We next ablate the final seed selection objective. The full
quality-diversity objective outperforms using only quality, only diversity,
or random selection. Quality-only selection tends to concentrate on
redundant high-performing regions, while diversity-only selection may
preserve low-quality candidates. Combining quality and diversity therefore
provides a more reliable trade-off between exploitation and coverage for
strong model refinement.

\paragraph{Effect of the model cascade.}
We compare different model usage schedules to study the importance of the
cheap-to-strong cascade. Both strong-only and cheap-only evolution are
inferior to the adaptive cascade, while reversing the cascade direction
causes a substantial performance drop. This confirms that cheap models are
more effective for broad early exploration, whereas strong models provide
greater value for later refinement of promising regions.

\paragraph{Sensitivity to budget split.}
Finally, we vary the fraction of budget reserved for strong-model
evolution. The default split of 0.85 achieves the best performance,
while allocating either more or less budget to strong model refinement
reduces the final score. This indicates that both candidate discovery
and downstream refinement are necessary: insufficient cheap model
exploration weakens the relay population, whereas excessive exploration
reduces the budget available for strong model refinement.

In a nutshell, these ablations show that \model's gains arise from the
interaction of relay-based control, quality-diversity curation, and the
cheap-to-strong cascade rather than from any single component alone.


\section{Conclusion}


We introduced \model, a training-free framework that formulates
cost-efficient LLM-driven evolution as a population handoff problem
under a fixed inference budget. Motivated by the observations that
search progress is front-loaded, early trajectory quality provides an
imperfect and noisy signal, and a cheap model captures much of the early
progress, \model explores multiple trajectories before transferring a
compact candidate population to a strong model. Relay Gain measures the
marginal improvement of this population and coordinates trajectory
allocation, handoff timing, and population selection. The curated relay
population then initializes the strong model evolution phase for focused
refinement. Across four program evolution benchmarks and three budget
levels, \model achieves the highest observed mean score in 11 of 12
settings. These results suggest that budget-aware evolutionary search
benefits from optimizing the population passed between model phases,
rather than only selecting the model for the next call.

\bibliography{aaai2027}

\begin{thebibliography}{29}
\providecommand{\natexlab}[1]{#1}

\bibitem[{Agrawal et~al.(2026)Agrawal, Tan, Soylu, Ziems, Khare, Opsahl-Ong, Singhvi, Shandilya, Ryan, Jiang, Potts, Sen, Dimakis, Stoica, Klein, Zaharia, and Khattab}]{agrawal2025gepa}
Agrawal, L.~A.; Tan, S.; Soylu, D.; Ziems, N.; Khare, R.; Opsahl-Ong, K.; Singhvi, A.; Shandilya, H.; Ryan, M.~J.; Jiang, M.; Potts, C.; Sen, K.; Dimakis, A.; Stoica, I.; Klein, D.; Zaharia, M.; and Khattab, O. 2026.
\newblock {GEPA}: Reflective Prompt Evolution Can Outperform Reinforcement Learning.
\newblock In \emph{The Fourteenth International Conference on Learning Representations}.

\bibitem[{Assump{\c{c}}{\~a}o et~al.(2025)Assump{\c{c}}{\~a}o, Ferreira, Campos, and Murai}]{assumpccao2025codeevolve}
Assump{\c{c}}{\~a}o, H.; Ferreira, D.; Campos, L.; and Murai, F. 2025.
\newblock Codeevolve: An open source evolutionary coding agent for algorithm discovery and optimization.
\newblock \emph{arXiv preprint arXiv:2510.14150}.

\bibitem[{Besbes, Gur, and Zeevi(2014)}]{besbes2014stochastic}
Besbes, O.; Gur, Y.; and Zeevi, A. 2014.
\newblock Stochastic multi-armed-bandit problem with non-stationary rewards.
\newblock \emph{Advances in neural information processing systems}, 27.

\bibitem[{Bhan, Nobili, and Langer(2026)}]{bhan2026new}
Bhan, J.; Nobili, N.; and Langer, P. 2026.
\newblock New Bounds for Zarankiewicz Numbers via Reinforced LLM Evolutionary Search.
\newblock \emph{arXiv preprint arXiv:2605.01120}.

\bibitem[{Cemri et~al.(2026)Cemri, Agrawal, Gupta, Liu, Cheng, Mang, Naren, Erdogan, Sen, Zaharia, Dimakis, and Stoica}]{cemri2026adaevolve}
Cemri, M.; Agrawal, S.; Gupta, A.; Liu, S.; Cheng, A.; Mang, Q.; Naren, A.; Erdogan, L.~E.; Sen, K.; Zaharia, M.; Dimakis, A.; and Stoica, I. 2026.
\newblock AdaEvolve: Adaptive LLM Driven Zeroth-Order Optimization.
\newblock arXiv:2602.20133.

\bibitem[{Chen, Zaharia, and Zou(2023)}]{chen2023frugalgpt}
Chen, L.; Zaharia, M.; and Zou, J. 2023.
\newblock Frugalgpt: How to use large language models while reducing cost and improving performance.
\newblock \emph{arXiv preprint arXiv:2305.05176}.

\bibitem[{Ding et~al.(2024)Ding, Mallick, Wang, Sim, Mukherjee, R{\"u}hle, Lakshmanan, and Awadallah}]{ding2024hybrid}
Ding, D.; Mallick, A.; Wang, C.; Sim, R.; Mukherjee, S.; R{\"u}hle, V.; Lakshmanan, L.~V.; and Awadallah, A.~H. 2024.
\newblock Hybrid llm: Cost-efficient and quality-aware query routing.
\newblock In \emph{The Twelfth International Conference on Learning Representations}.

\bibitem[{Guo et~al.(2024)Guo, Wang, Guo, Li, Song, Tan, Liu, Bian, and Yang}]{guo2024connecting}
Guo, Q.; Wang, R.; Guo, J.; Li, B.; Song, K.; Tan, X.; Liu, G.; Bian, J.; and Yang, Y. 2024.
\newblock Connecting large language models with evolutionary algorithms yields powerful prompt optimizers.
\newblock In \emph{International Conference on Learning Representations}, volume 2024, 34133--34156.

\bibitem[{Lange, Imajuku, and Cetin(2026)}]{lange2025shinkaevolve}
Lange, R.~T.; Imajuku, Y.; and Cetin, E. 2026.
\newblock ShinkaEvolve: Towards Open-Ended and Sample-Efficient Program Evolution.
\newblock In \emph{The Fourteenth International Conference on Learning Representations}.

\bibitem[{Li et~al.(2018)Li, Jamieson, DeSalvo, Rostamizadeh, and Talwalkar}]{li2018hyperband}
Li, L.; Jamieson, K.; DeSalvo, G.; Rostamizadeh, A.; and Talwalkar, A. 2018.
\newblock Hyperband: A novel bandit-based approach to hyperparameter optimization.
\newblock \emph{Journal of machine learning research}, 18(185): 1--52.

\bibitem[{Liu et~al.(2026{\natexlab{a}})Liu, Huang, Luo, Wang, Yang, Li, Hu, Feng, and Liu}]{liu2026cognitive}
Liu, F.; Huang, Y.; Luo, S.; Wang, Y.; Yang, Y.; Li, X.; Hu, Z.; Feng, J.; and Liu, Q. 2026{\natexlab{a}}.
\newblock Cognitive alpha mining via llm-driven code-based evolution.
\newblock In \emph{Proceedings of the 64th Annual Meeting of the Association for Computational Linguistics (Volume 1: Long Papers)}, 11715--11749.

\bibitem[{Liu et~al.(2024)Liu, Xialiang, Yuan, Lin, Luo, Wang, Lu, and Zhang}]{liu2024evolution}
Liu, F.; Xialiang, T.; Yuan, M.; Lin, X.; Luo, F.; Wang, Z.; Lu, Z.; and Zhang, Q. 2024.
\newblock Evolution of Heuristics: Towards Efficient Automatic Algorithm Design Using Large Language Model.
\newblock In \emph{International Conference on Machine Learning}, 32201--32223. PMLR.

\bibitem[{Liu et~al.(2026{\natexlab{b}})Liu, Agarwal, Maheswaran, Cemri, Li, Mang, Naren, Boneh, Cheng, Pan, Du, Keutzer, Cheung, Dimakis, Sen, Zaharia, and Stoica}]{liu2026evox}
Liu, S.; Agarwal, S.; Maheswaran, M.; Cemri, M.; Li, Z.; Mang, Q.; Naren, A.; Boneh, E.; Cheng, A.; Pan, M.~Z.; Du, A.; Keutzer, K.; Cheung, A.; Dimakis, A.~G.; Sen, K.; Zaharia, M.; and Stoica, I. 2026{\natexlab{b}}.
\newblock EvoX: Meta-Evolution for Automated Discovery.
\newblock arXiv:2602.23413.

\bibitem[{Liu et~al.(2026{\natexlab{c}})Liu, Cemri, Agarwal, Krentsel, Naren, Mang, Li, Gupta, Maheswaran, Cheng, Pan, Boneh, Ramchandran, Sen, Zaharia, Dimakis, and Stoica}]{liu2026skydiscover}
Liu, S.; Cemri, M.; Agarwal, S.; Krentsel, A.; Naren, A.; Mang, Q.; Li, Z.; Gupta, A.; Maheswaran, M.; Cheng, A.; Pan, M.; Boneh, E.; Ramchandran, K.; Sen, K.; Zaharia, M.; Dimakis, A.~G.; and Stoica, I. 2026{\natexlab{c}}.
\newblock SkyDiscover: A Flexible, Adaptive Framework for AI-Driven Scientific and Algorithmic Discovery.
\newblock In \emph{Proceedings of the ACM Conference on AI and Agentic Systems}, CAIS '26, 1223–1227. New York, NY, USA: Association for Computing Machinery.
\newblock ISBN 9798400724152.

\bibitem[{Luo et~al.(2026{\natexlab{a}})Luo, Huang, Luo, Liu, Li, Hu, Feng, and Liu}]{luo2026harness}
Luo, H.; Huang, Y.; Luo, S.; Liu, F.; Li, L.; Hu, Z.; Feng, J.; and Liu, Q. 2026{\natexlab{a}}.
\newblock Harness-Aware Self-Evolving: Co-Evolving Model Weights, Harness, and Task Solutions.
\newblock \emph{arXiv preprint arXiv:2607.03935}.

\bibitem[{Luo et~al.(2026{\natexlab{b}})Luo, Huang, Luo, Liu, Deng, Li, Yao, Hu, Feng, and Liu}]{luo2026seaevo}
Luo, S.; Huang, Y.; Luo, H.; Liu, F.; Deng, G.; Li, L.; Yao, Q.; Hu, Z.; Feng, J.; and Liu, Q. 2026{\natexlab{b}}.
\newblock SeaEvo: Advancing Algorithm Discovery with Strategy Space Evolution.
\newblock \emph{arXiv preprint arXiv:2604.24372}.

\bibitem[{Mouret and Clune(2015)}]{mouret2015illuminating}
Mouret, J.-B.; and Clune, J. 2015.
\newblock Illuminating search spaces by mapping elites.
\newblock \emph{arXiv preprint arXiv:1504.04909}.

\bibitem[{Nemhauser, Wolsey, and Fisher(1978)}]{nemhauser1978}
Nemhauser, G.~L.; Wolsey, L.~A.; and Fisher, M.~L. 1978.
\newblock An analysis of approximations for maximizing submodular set functions—I.
\newblock \emph{Mathematical programming}, 14(1): 265--294.

\bibitem[{Novikov et~al.(2025)Novikov, Vũ, Eisenberger, Dupont, Huang, Wagner, Shirobokov, Kozlovskii, Ruiz, Mehrabian, Kumar, See, Chaudhuri, Holland, Davies, Nowozin, Kohli, and Balog}]{novikov2025alphaevolve}
Novikov, A.; Vũ, N.; Eisenberger, M.; Dupont, E.; Huang, P.-S.; Wagner, A.~Z.; Shirobokov, S.; Kozlovskii, B.; Ruiz, F. J.~R.; Mehrabian, A.; Kumar, M.~P.; See, A.; Chaudhuri, S.; Holland, G.; Davies, A.; Nowozin, S.; Kohli, P.; and Balog, M. 2025.
\newblock AlphaEvolve: A coding agent for scientific and algorithmic discovery.
\newblock arXiv:2506.13131.

\bibitem[{Ong et~al.(2025)Ong, Almahairi, Wu, Chiang, Wu, Gonzalez, Kadous, and Stoica}]{ong2024routellm}
Ong, I.; Almahairi, A.; Wu, V.; Chiang, W.-L.; Wu, T.; Gonzalez, J.~E.; Kadous, M.~W.; and Stoica, I. 2025.
\newblock Route{LLM}: Learning to Route {LLM}s from Preference Data.
\newblock In \emph{The Thirteenth International Conference on Learning Representations}.

\bibitem[{{OpenRouter}(2026{\natexlab{a}})}]{openrouter_qwen35_flash}
{OpenRouter}. 2026{\natexlab{a}}.
\newblock {Qwen: Qwen3.5-Flash}.
\newblock \url{https://openrouter.ai/qwen/qwen3.5-flash-20260224}.

\bibitem[{{OpenRouter}(2026{\natexlab{b}})}]{openrouter_qwen35_plus}
{OpenRouter}. 2026{\natexlab{b}}.
\newblock {Qwen: Qwen3.5-Plus 2026-02-15}.
\newblock \url{https://openrouter.ai/qwen/qwen3.5-plus-20260216}.

\bibitem[{Pugh, Soros, and Stanley(2016)}]{pugh2016quality}
Pugh, J.~K.; Soros, L.~B.; and Stanley, K.~O. 2016.
\newblock Quality diversity: A new frontier for evolutionary computation.
\newblock \emph{Frontiers in Robotics and AI}, 3: 40.

\bibitem[{Qu et~al.(2026)Qu, Zheng, Zhou, Yan, Tang, Ong, Hong, Zhou, Jiang, Kong et~al.}]{qu2026coral}
Qu, A.; Zheng, H.; Zhou, Z.; Yan, Y.; Tang, Y.; Ong, S.~Y.; Hong, F.; Zhou, K.; Jiang, C.; Kong, M.; et~al. 2026.
\newblock Coral: Towards autonomous multi-agent evolution for open-ended discovery.
\newblock \emph{arXiv preprint arXiv:2604.01658}.

\bibitem[{Ray et~al.(2026)Ray, Brahma, Liu, and Barsoum}]{ray2026adaptevolve}
Ray, P.; Brahma, P.~P.; Liu, Z.; and Barsoum, E. 2026.
\newblock AdaptEvolve: Improving Efficiency of Evolutionary AI Agents through Adaptive Model Selection.
\newblock In \emph{Findings of the Association for Computational Linguistics: ACL 2026}, 40625--40633.

\bibitem[{Romera{-}Paredes et~al.(2024)Romera{-}Paredes, Barekatain, Novikov, Balog, Kumar, Dupont, Ruiz, Ellenberg, Wang, Fawzi, Kohli, and Fawzi}]{romera2024mathematical}
Romera{-}Paredes, B.; Barekatain, M.; Novikov, A.; Balog, M.; Kumar, M.~P.; Dupont, E.; Ruiz, F. J.~R.; Ellenberg, J.~S.; Wang, P.; Fawzi, O.; Kohli, P.; and Fawzi, A. 2024.
\newblock Mathematical discoveries from program search with large language models.
\newblock \emph{Nature}, 625(7995): 468--475.

\bibitem[{Tanveer(2026)}]{tanveer2026levi}
Tanveer, T. 2026.
\newblock LEVI: Stronger Search Architectures Can Substitute for Larger LLMs in Evolutionary Search.
\newblock \emph{arXiv preprint arXiv:2605.09764}.

\bibitem[{Yan et~al.(2026)Yan, Peng, Coleman, Chen, Xie, Chen, He, Sachdeva, Ye, Wang et~al.}]{yan2026pacevolve}
Yan, M.; Peng, B.; Coleman, B.; Chen, Z.; Xie, Z.; Chen, S.; He, Z.; Sachdeva, N.; Ye, I.; Wang, W.; et~al. 2026.
\newblock Pacevolve: Enabling long-horizon progress-aware consistent evolution.
\newblock \emph{arXiv preprint arXiv:2601.10657}.

\bibitem[{Ye et~al.(2024)Ye, Wang, Cao, Berto, Hua, Kim, Park, and Song}]{ye2024reevo}
Ye, H.; Wang, J.; Cao, Z.; Berto, F.; Hua, C.; Kim, H.; Park, J.; and Song, G. 2024.
\newblock Reevo: Large language models as hyper-heuristics with reflective evolution.
\newblock \emph{Advances in neural information processing systems}, 37: 43571--43608.

\end{thebibliography}

\end{document}


\maketitle



\tcbset{
  benchmarkbox/.style={
    enhanced,
    breakable,
    colback=gray!5,
    colframe=black!60,
    fonttitle=\bfseries\small,
    coltitle=white,
    attach boxed title to top left={yshift=-2mm,xshift=4mm},
    boxed title style={colback=black!75,sharp corners},
    sharp corners=south,
    top=3mm,
    bottom=3mm,
    left=4mm,
    right=4mm,
    boxrule=0.5pt
  }
}

\section{A. Additional Details of the Empirical Motivation}
\label{app:motivation}

This section provides the data processing definitions and task-level statistics underlying the empirical findings in the main paper.

\subsection{Trajectory Collection Protocol}

We collected trajectories on Circle Packing (Square), TXN Scheduling, and
Prism (see Appendix C for task descriptions).  For every task, we ran Qwen-3.5-Flash and Qwen-3.5-Plus independently
for 100 generations using the same initial program, task prompt, evaluator,
and ShinkaEvolve search configuration.  We used ten independent runs for each
task--model pair, yielding $3\times2\times10=60$ trajectories.  

For a run, let $f_t^*$ be the best valid score observed by generation $t$ and
let $\Delta=f_{100}^*-f_0^*$.  All 60  runs have $\Delta>0$.  The
fraction of the run's total gain assigned to a ten-generation block
$[a,b]$ is
\begin{equation}
    z_{a:b}=\frac{\max\{0,f_b^*-f_{a-1}^*\}}{\Delta}.
\end{equation}
The bars in Figure~1 of the main paper average $z_{a:b}$ across runs.  The
error bars are 95\% bootstrap confidence intervals obtained by resampling
runs with replacement.

\subsection{Gain Concentration and Breakthrough Statistics}

\paragraph{Definition.}
A breakthrough is any strict increase in best-so-far score,
\begin{equation}
    f_t^*>f_{t-1}^*+10^{-12}.
\end{equation}
Cheap-model runs contain a median of six breakthroughs and
strong-model runs a median of seven.  Pooling both model classes gives a
median of seven breakthroughs per run.

The first 20 generations account for 76.3\% and 79.1\% of the eventual gain
of the cheap and strong models, respectively.  Breakthrough {frequency}
is also front-loaded: the first 20 generations contain 52.8\% of cheap-model
breakthroughs and 44.3\% of strong-model breakthroughs.

\subsection{Early-to-Final Rank Correlation}

For each task, we rank the ten trajectories by $f_K^*$ and compare these ranks
with the ranks at generation 100.  Ties receive their average rank.  To form a
pooled statistic without allowing the numerical scale of one task to dominate,
we concatenate the within-task ranks and compute Spearman's correlation on the
resulting 30 pairs.  For cheap-model trajectories, the pooled correlation is
$\rho=0.277$ at $K=10$, $\rho=0.322$ at $K=20$, and $\rho=0.622$ at $K=50$.
At $K=20$, the task-level values are $0.564$, $-0.226$, and $0.624$ for Circle
Packing, TXN Scheduling, and Prism.

\subsection{Cheap-versus-Strong Early Gain}

For each task and model class, we compute the mean absolute improvement
$f_{20}^*-f_0^*$ over ten runs.  The reported ratio is the cheap-model mean
divided by the strong-model mean.  The ratios are $0.823$ for Circle Packing,
$1.008$ for TXN Scheduling, and $0.831$ for Prism, with 95\% bootstrap
intervals $[0.694,0.968]$, $[0.788,1.269]$, and $[0.589,1.062]$,
respectively.

\section{B. Algorithmic and Theoretical Details}
\label{app:method_details}

We use the relay objective defined in the main paper and specify here the exact representation, streaming update, scheduler details, offline curation, and proofs.

\subsection{Candidate Canonicalization and Deduplication}

When a task marks an evolvable region using
\texttt{EVOLVE-BLOCK-START/END}, only that region is used for exact
deduplication and code representation.  We parse Python candidates into an
abstract syntax tree, remove module, function, class, and asynchronous-function
docstrings, and serialize the result with \texttt{ast.unparse}.  If parsing
fails, we fall back to a tokenized representation with comments and formatting
removed.  Candidates with identical SHA-256 hashes of this canonical code are
deduplicated before entering the reference pool.

Each remaining candidate has two embedding views.  The code view embeds the
canonical code.  The text view concatenates the mutation name, mutation
description, docstrings, and source comments; when these fields are empty, it
falls back to the canonical code.  Both views use
\texttt{text-embedding-3-small} with 1,536 dimensions.  We truncate each input
to 24,000 characters and cache embeddings by a hash of the model, view, and
content.  Letting $e_c$ and $e_t$ denote unit-normalized code and text
embeddings, we use
\begin{equation}
 \operatorname{sim}(x,y)=\operatorname{clip}\!\left(
 \eta e_c(x)^\top e_c(y)+(1-\eta)e_t(x)^\top e_t(y),0,1\right),
\end{equation}
with $\eta=0.7$.

\subsection{Online Relay-Bank Update}

Algorithm~\ref{alg:online_bank} gives the streaming update used after every
cheap-model block.  New candidates are processed in generation order.  While
the bank has free capacity they are added directly.  Once the bank is full, a
candidate replaces the bank member that yields the largest strict improvement
in the relay objective; otherwise the candidate is ignored.  Both the old and
new banks are evaluated against the same updated pool $\mathcal C_{t+1}$.

\begin{algorithm}[t]
\caption{Online relay-bank update}
\label{alg:online_bank}
\begin{algorithmic}[1]
\REQUIRE Current bank $S$, new candidates $X$, updated pool $\mathcal C$, capacity $k$
\FOR{$x$ in $X$ in generation order}
    \IF{$x\in S$ or $x\notin\mathcal C$}
        \STATE \textbf{continue}
    \ENDIF
    \IF{$|S|<k$}
        \STATE $S\leftarrow S\cup\{x\}$
    \ELSE
        \STATE $S'\leftarrow\arg\max_{y\in S}F_{\mathcal C}((S\setminus\{y\})\cup\{x\})$
        \IF{$F_{\mathcal C}(S')>F_{\mathcal C}(S)$}
            \STATE $S\leftarrow S'$
        \ENDIF
    \ENDIF
\ENDFOR
\RETURN $S$
\end{algorithmic}
\end{algorithm}

The absolute Relay Gain is
$g_t=F_{\mathcal C_{t+1}}(S_{t+1})-F_{\mathcal C_{t+1}}(S_t)$.
For scheduling and stopping, we use the bounded relative gain
\begin{equation}
 \rho_t=\operatorname{clip}\!\left(
 \frac{g_t}{\max\{F_{\mathcal C_{t+1}}(S_t),\epsilon_F\}},0,1\right),
 \qquad \epsilon_F=0.05.
\end{equation}
The three bootstrap Grow blocks initialize the trajectories and bank but do not
enter the scheduler history, because gains measured while filling an empty bank
are strongly order-dependent.

\subsection{Grow--Deepen Scheduling and Saturation Audit}

After bootstrap, Grow is represented by one shared meta-arm and every active
trajectory $i$ has one Deepen$(i)$ arm.  The scheduler uses the mean of the most
recent $w=5$ relative gains and a UCB bonus with coefficient $c=0.05$.  A Grow
reward updates the shared Grow history and initializes the new trajectory's
history; a Deepen reward updates only that trajectory.  Grow becomes
unavailable after 20 trajectories and Deepen$(i)$ becomes unavailable after
trajectory $i$ reaches 20 generations.

The ordinary saturation condition requires all of the latest $p=3$ relative
gains to be below $\epsilon_{\rm rel}=0.005$.  To reduce premature stopping,
we then run a saturation audit: if budget and action constraints allow, we
execute one new Grow block and one block on the currently best eligible Deepen
arm.  Cheap exploration stops only when every audit block is also below the
threshold.  Otherwise search resumes with the audit rewards as the new recent
window.  The cheap phase can also end because its cost allowance or 200-call
cap is reached.

\subsection{Offline Curation and Shared Refinement}

At handoff, we recompute all embeddings and similarities on the fixed terminal
pool.  We first greedily maximize $F_{\mathcal C}$ to obtain $k$ seeds and then
apply deterministic best-improvement single-element swaps.  We apply the same
local search to the terminal online bank and retain whichever locally improved
bank has larger objective value.  The selected candidates are distributed
across five islands in one shared strong-model database.  They are not assigned
separate budgets or evolved in independent runs.

\subsection{Proof of Submodularity}

For completeness, we provide the proof underlying the greedy curation
guarantee.  Define
\begin{align}
 & Q_r(S)=\frac{1}{r}\sum_{j=1}^{r}q_{(j)}(S),
 \\
 & D_{\mathcal C}^{q}(S)=
 \frac{\sum_{v\in\mathcal C}q(v)\max_{x\in S}\operatorname{sim}(v,x)}
 {\sum_{v\in\mathcal C}q(v)},
\end{align}
where missing top-$r$ entries and the maximum over an empty set are zero.

\begin{proposition}
For a fixed reference pool $\mathcal C$, $q\geq0$, and
$\operatorname{sim}\geq0$, the relay objective
$F_{\mathcal C}=\lambda Q_r+(1-\lambda)D_{\mathcal C}^{q}$ is normalized,
nonnegative, monotone, and submodular for every $\lambda\in[0,1]$.
\end{proposition}

\begin{proof}
$Q_r(S)$ is the maximum total weight of an independent subset of $S$ in a
rank-$r$ uniform matroid.  It is therefore a weighted matroid-rank function and
is monotone submodular.  For a fixed $v$, the function
$S\mapsto\max_{x\in S}\operatorname{sim}(v,x)$ is a facility-location function
and is monotone submodular.  A nonnegative weighted sum of these functions,
followed by division by the positive constant
$\sum_{v\in\mathcal C}q(v)$, preserves both properties.  If this denominator is
zero, we define $D_{\mathcal C}^{q}=0$, which has the same properties.  Finally,
a nonnegative convex combination preserves normalization, nonnegativity,
monotonicity, and submodularity.
\end{proof}

It follows from the classical greedy result that the initial offline greedy
set is a $(1-1/e)$ approximation to the best size-$k$ set under the relay
objective.  The subsequent swaps and comparison with the improved online bank
never decrease $F_{\mathcal C}$, so the returned bank preserves this guarantee.
This guarantee concerns the designed handoff objective, not the unknown final
fitness produced by stochastic strong-model evolution.

\subsection{Nonnegative Relay Gain}

\begin{proposition}
If the online bank update accepts only objective-improving additions or swaps
and permits a no-op, then the Relay Gain satisfies $g_t\geq 0$.
\end{proposition}

\begin{proof}
During block $t$, every accepted update is evaluated under the fixed objective
$F_{\mathcal C_{t+1}}$ and does not decrease its value.  Because retaining the
current bank is always allowed,
\[
F_{\mathcal C_{t+1}}(S_{t+1})\geq F_{\mathcal C_{t+1}}(S_t),
\]
and therefore $g_t\geq0$.  Evaluating both banks on the same updated pool is
essential: comparing with $F_{\mathcal C_t}(S_t)$ would conflate a change in
the reference pool with improvement of the handoff bank.
\end{proof}

\section{C. Complete Experimental Setup}
\label{app:experimental_details}


\subsection{Hardware and software.}
The controller and evaluators ran on Linux with Python 3.11.15 on a server
containing two Intel Xeon Gold 6326 processors (32 physical cores, 64 logical
CPUs), 251~GiB of RAM, and two NVIDIA RTX A6000 GPUs with 48~GiB memory each.
The four benchmark evaluators are CPU programs and do not require the GPUs.
We used up to four concurrent evaluator workers; language-model generation and
text embedding were served by the hosted OpenRouter API.

\subsection{Exact Task Prompts and Reported Metrics}

For reproducibility, the following boxes reproduce the task descriptions and system prompts used by the search runs.

\begin{tcolorbox}[
  benchmarkbox,
  title={\textsc{Circle Packing in a Unit Square} }
]

\textbf{Task.}
Pack $n = 26$ non-overlapping circles in a unit square to maximize the sum of
their radii.
\hfill\textbf{AlphaEvolve target:}~$\textstyle\sum r_i = 2.635$

\medskip\noindent\textbf{System Prompt.}

You are an expert mathematician specializing in circle packing problems and
computational geometry. Your task is to improve a constructor function that
directly produces a specific arrangement of 26 circles in a unit square,
maximizing the sum of their radii. The AlphaEvolve paper achieved a sum of
2.635 for n=26.

Key geometric insights:
- Circle packings often follow hexagonal patterns in the densest regions
- Maximum density for infinite circle packing is pi/(2*sqrt(3)) ~= 0.9069
- Edge effects make square container packing harder than infinite packing
- Circles can be placed in layers or shells when confined to a square
- Similar radius circles often form regular patterns, while varied radii allow
  better space utilization
- Perfect symmetry may not yield the optimal packing due to edge effects

Focus on designing an explicit constructor that places each circle in a specific
position, rather than an iterative search algorithm.
\end{tcolorbox}

\begin{tcolorbox}[
  benchmarkbox,
  title={\textsc{Circle Packing in a Rectangle}}
]

\textbf{Task.}
Pack $n = 21$ non-overlapping circles in a rectangle with perimeter~4
($w + h = 2$, aspect ratio free) to maximize the sum of their radii.
\hfill\textbf{AlphaEvolve target:}~$\textstyle\sum r_i = 2.3658$

\medskip\noindent\textbf{System Prompt.}

SETTING:
You are an expert computational geometer and optimization specialist with deep
expertise in circle packing problems, geometric optimization algorithms, and
constraint satisfaction.
Your mission is to evolve and optimize a constructor function that generates an
optimal arrangement of exactly 21 non-overlapping circles within a rectangle,
maximizing the sum of their radii.

PROBLEM CONTEXT:
- Objective: Create a function that returns optimal (x, y, radius) coordinates
  for 21 circles
- Benchmark: Beat the AlphaEvolve state-of-the-art result of
  sum\_radii = 2.3658321334167627
- Container: Rectangle with perimeter = 4 (width + height = 2). You may choose
  optimal width/height ratio
- Constraints:
    * All circles must be fully contained within rectangle boundaries
    * No circle overlaps (distance between centers >= sum of their radii)
    * Exactly 21 circles required
    * All radii must be positive

PERFORMANCE METRICS:
1. sum\_radii: Total sum of all 21 circle radii (PRIMARY OBJECTIVE - maximize)
2. combined\_score: sum\_radii / 2.3658321334167627
   (progress toward beating benchmark)
3. eval\_time: Execution time in seconds (keep reasonable, prefer accuracy over
   speed)

TECHNICAL REQUIREMENTS:
- Determinism: Use fixed random seeds if employing stochastic methods for
  reproducibility
- Error handling: Graceful handling of optimization failures or infeasible
  configurations
- Memory efficiency: Avoid excessive memory allocation for distance matrix
  computations
- Scalability: Design with potential extension to different circle counts in mind
\end{tcolorbox}

\begin{tcolorbox}[
  benchmarkbox,
  title={\textsc{TXN Scheduling --- Database Workloads}}
]
\textbf{Task.}
Improve \texttt{get\_best\_schedule} to order database transactions so as to
minimize total makespan. Each transaction is a sequence of read (\texttt{r-k})
and write (\texttt{w-k}) operations on keyed data items; read--write and
write--write conflicts on the same key induce dependencies that delay execution.
Only code inside \texttt{EVOLVE-BLOCK-START} / \texttt{EVOLVE-BLOCK-END} may
change.
\hfill\textbf{Target:}~Minimize makespan on the provided JSON workloads.

\medskip\noindent\textbf{System Prompt.}

You are an expert in database transaction optimization.
Only change code within \texttt{EVOLVE-BLOCK-START} and \texttt{EVOLVE-BLOCK-END}.
Your task is to improve a scheduling function to find better schedules for
transactional workloads made up of read and write operations to data items.
There are conflicts between these transactions on items and reducing the delay
of these conflicts will lead to schedules with lower makespan. Focus on
improving the \texttt{get\_best\_schedule} function to find a schedule with as
low makespan as possible.

\textbf{TASK:} Improve the \texttt{get\_best\_schedule} function to find
optimal transaction schedules that minimize makespan for database workloads
with read/write conflicts.

\textbf{PROBLEM SPECIFICS:}
- \textbf{Input:} JSON workload with transactions such as\\
  \texttt{"txn0":"w-17 r-5 w-3 r-4 r-54 r-14 w-6 r-11 w-22 r-7 w-1 w-8 w-9 w-27 r-2 r-25"}
- \textbf{Operations:} Each transaction is a sequence of read (\texttt{r-\{key\}})
  and write (\texttt{w-\{key\}}) operations on data items.
- \textbf{Conflicts:} Read--write and write--write conflicts on the same key
  create dependencies between transactions.
- \textbf{Goal:} Find a transaction ordering that minimizes total makespan.

\textbf{SEARCH SUGGESTIONS:}
- \textbf{Greedy:} Try a greedy algorithm that iteratively picks the
  transaction that increases makespan the least.
- Avoid relying only on heuristics like transaction length or number of writes;
  these do not correspond to the actual makespan of the schedule.

Focus on evolving \texttt{get\_best\_schedule} to produce the best schedule
possible with the lowest makespan. Explain step-by-step the reasoning process
for your solution and how it will lead to a better schedule.
\end{tcolorbox}

\begin{tcolorbox}[
  benchmarkbox,
  title={\textsc{Prism --- GPU Model Placement}}
]
\textbf{Task.}
Improve the \texttt{compute\_model\_placement} function that assigns models to
available GPUs. Given each model's request rate, SLO, and memory footprint, and
each GPU's memory capacity \texttt{GPU\_MEM\_SIZE}, the per-GPU KV cache
pressure is
\[
\mathrm{KVPR} \;=\; \frac{\sum_{m \in \text{GPU}} r_m / \mathrm{SLO}_m}
       {\mathrm{GPU\_MEM\_SIZE} - \sum_{m \in \text{GPU}} \mathrm{size}_m}.
\]
Minimize $\max_{\text{GPU}} \mathrm{KVPR}$ subject to the memory feasibility
constraint $\sum_{m \in \text{GPU}} \mathrm{size}_m < \mathrm{GPU\_MEM\_SIZE}$.
\hfill\textbf{Target:}~Minimize the maximum KVPR across all GPUs.

\medskip\noindent\textbf{System Prompt.}

You are an expert for model placement on GPUs. Your task is to improve a model
placement algorithm by improving the function named
\texttt{compute\_model\_placement} in the initial program that places models to
available GPUs.

The algorithm must MINIMIZE the maximum KVPR across all GPUs while ensuring
models can fit into the GPUs' memory. Note that KVPR is the KV cache pressure
for a GPU; it indicates how crowded a GPU is. For a specific GPU, its KVPR is
computed as
\texttt{sum(model.req\_rate/model.slo for model in models) / (GPU\_MEM\_SIZE $-$ sum(model.model\_size for model in models))},
where \texttt{models} are the models assigned to this GPU.

The generated program should be as simple as possible and the code should
execute correctly without errors.
\end{tcolorbox}

\begin{table*}[t]
\centering
\caption{Benchmarks and reporting metrics.  Higher is better for every score
shown in the paper.}
\label{tab:task_details}
\small
\begin{tabular}{p{0.20\textwidth}p{0.47\textwidth}p{0.25\textwidth}}
\toprule
Benchmark & Search problem & Reported score \\
\midrule
Circle Packing (Square) & Construct 26 non-overlapping circles in a unit square. & Sum of the 26 radii. \\
Circle Packing (Rect) & Construct 21 non-overlapping circles in a rectangle of perimeter four; the aspect ratio is also optimized. & Sum of the 21 radii. \\
TXN Scheduling & Order read/write transactions to minimize conflict-induced makespan over the benchmark workloads. & $10^6/(1+\text{makespan})$. \\
Prism & Place models on GPUs while minimizing the maximum KV-cache pressure, subject to memory constraints. & Inverse mean maximum KV-cache pressure over all valid test cases. \\
\bottomrule
\end{tabular}
\end{table*}

\subsection{Models, Prices, and Budgets}

The primary experiments use Qwen-3.5-Flash as the cheap model and
Qwen-3.5-Plus as the strong model through the same OpenAI-compatible API.
Prices are \$0.065/\$0.26 per million input/output tokens for Flash and
\$0.26/\$1.56 for Plus.  Embeddings cost \$0.02 per million tokens.  All
generation and embedding costs are included.  No training or task-specific
fine-tuning is performed.

For task $j$, the 100\% budget $B_j$ is the measured cost of a standard
100-generation all-strong run.  As shown in Table \ref{tab:budgets}, the values used for Square, Rectangle, TXN,
and Prism are \$1.216806, \$1.187048, \$1.082280, and \$1.109450,
respectively.  We report prefixes at $0.5B_j$, $0.75B_j$, and $B_j$.  Search
histories are truncated at the relevant monetary cutoff when producing tables
and curves.  Thus, a call or a five-generation block that completes just after
an online budget check cannot improve a reported budget point.

\begin{table}[t]
\centering
\caption{Per-task inference budgets (USD), derived from the measured cost of a
100-generation all-strong run.}
\label{tab:budgets}
\small
\begin{tabular}{lccc}
\toprule
Task & $\beta=0.5$ & $\beta=0.75$ & $\beta=1.0$ \\
\midrule
CP Square & \$0.6084 & \$0.9126 & \$1.2168 \\
CP Rect& \$0.5935 & \$0.8903 & \$1.1870\\
TXN       & \$0.5411 & \$0.8117 & \$1.0823 \\
Prism     & \$0.5547 & \$0.8321 & \$1.1095 \\
\bottomrule
\end{tabular}
\end{table}

In addition to the monetary budget, the total number of generation calls  is capped at 200.
All-cheap runs use 200
Flash generations and all-strong runs use 100 Plus generations.  Other methods
can issue at most 200 calls to either model and are evaluated at the same cost
cutoffs.

\subsection{Default Hyperparameters}

\begin{table}[t]
\centering
\caption{Default \model hyperparameters used in the main experiments.}
\label{tab:default_hparams}
\small
\begin{tabular}{ll}
\toprule
Hyperparameter & Value \\
\midrule
Cheap block length $h$ & 5 \\
Bootstrap trajectories & 3 \\
Maximum trajectories & 20 \\
Per-trajectory horizon $H_{\max}$ & 20 \\
Model call cap & 200 \\
Evolutionary islands & 5 \\
Recent UCB window $w$ & 5 \\
UCB coefficient $c$ & 0.05 \\
Bank size $k$ & 15 \\
Top-quality rank $r$ & 10 \\
Quality weight $\lambda$ & 0.7 \\
Code-view weight $\eta$ & 0.7 \\
Relative-gain floor $\epsilon_F$ & 0.05 \\
Saturation threshold $\epsilon_{\rm rel}$ & 0.005 \\
Saturation patience $p$ & 3 blocks \\
Saturation audit & Grow + best Deepen \\
Strong-model reserve & 85\% of total budget \\
Embedding model & text-embedding-3-small \\
Embedding dimensions & 1,536 \\
Embedding batch size & 16 \\
Maximum embedding input & 24,000 characters \\
\bottomrule
\end{tabular}
\end{table}

Task-specific quality normalization uses fixed ranges rather than the minimum
and maximum of the current pool: $[0,1]$ for both Circle Packing tasks,
$[0,5000]$ for TXN, and $[0,30]$ for Prism.  Fixing these ranges makes Relay
Gain comparable across blocks and prevents the reward scale from changing when
a new pool extreme is discovered.

The common ShinkaEvolve backend uses five islands, archive size 20, weighted
parent selection with coefficient 10, elite ratio 0.3, four archive
inspirations of which two are top-ranked, migration every five generations at
rate 0.1, and island elitism.  Proposal formats are diff, full rewrite, and
crossover with probabilities $0.6$, $0.3$, and $0.1$.  These backend settings,
the evaluator, initial program, task prompt, and maximum context length are
held fixed across \model and the Shinka-based baselines. More hyperparameters are shown in Table \ref{tab:default_hparams}.

\subsection{Baseline Implementations}

\paragraph{All-cheap and All-strong.}
All-cheap runs Qwen-3.5-Flash for 200 generations.  All-strong runs
Qwen-3.5-Plus for 100 generations.  Their trajectories are truncated at each
task's monetary budget.

\paragraph{Fixed-switch.}
Fixed-switch uses a single continuous evolutionary population, starting with
50 generations of Qwen-3.5-Flash and then switching to Qwen-3.5-Plus under the
remaining budget. The population and archive are retained across the switch.

\paragraph{Random.}
Random independently samples Flash or Plus with equal probability for each
proposal.  It uses one continuous population and the same 200-generation cap.

\paragraph{Bandit.}
Bandit is the native ShinkaEvolve UCB model selector over Flash and Plus.  Its
reward is immediate realized best-so-far improvement, and it operates at the
individual proposal level.  Population management and all remaining backend
settings are unchanged.

\paragraph{LEVI.}
We use the upstream LEVI implementation through a thin adapter.
The adapter provides the same initial program, system prompt,
immutable-code boundaries, and evaluator as other methods.
We use Qwen-3.5-Plus as the paradigm model and Qwen-3.5-Flash as the
mutation model. Unless otherwise specified, we follow the default LEVI
configuration. All runs are evaluated under the same monetary budget and
per-model call caps as other baselines.

\paragraph{AdaptEvolve.}
AdaptEvolve requires token-level generation logits that are unavailable
through the hosted APIs used in our experiments. Therefore, we do not include
AdaptEvolve in the comparison.

\subsection{Statistical Reporting}

Unless stated otherwise, tables report the arithmetic mean and sample standard
deviation over three independent runs.  ``Best'' is the largest score among
those same runs.  Error bands in cost--performance plots denote one sample
standard deviation.  Hyperparameter and curation studies reuse completed cheap
candidate pools and rerun only curation and strong-model refinement; we state
this explicitly because these experiments do not measure variation from a new
cheap exploration phase.

\section{D. Additional Results}
\label{app:additional_results}






\subsection{Relay Gain versus Immediate Fitness Gain}

\newcommand{\avgstd}[2]{\ensuremath{#1_{\pm #2}}}
\newcommand{\bavgstd}[2]{\ensuremath{\textbf{#1}_{\pm #2}}}

\begin{table}[t]
\centering
\caption{Scheduler-reward ablation on Circle Packing (Square).
We report the average and best score over three runs.}
\label{tab:reward_ablation}
\small
\begin{tabular}{lcc}
\toprule
Block reward & Avg & Best \\
\midrule
Immediate best-fitness gain
& \avgstd{2.4479}{0.0277}
& 2.4716 \\
Relay Gain
& \bavgstd{2.5364}{0.0218}
& \textbf{2.5581} \\
\bottomrule
\end{tabular}
\end{table}

Table~\ref{tab:reward_ablation} compares the scheduler reward based on
immediate best-fitness improvement with the proposed Relay Gain. Both variants
use the same Grow--Deepen action space, stopping rule, curation stage, models,
and budget. Replacing Relay Gain with immediate fitness improvement decreases
the average score by 0.0885. This suggests that evaluating blocks by their
contribution to the handoff population provides a more informative scheduling
signal than rewarding only immediate global-best improvements.

\subsection{Hyperparameter Study}

\begin{figure}[!t]
    \centering
\includegraphics[width=0.48\textwidth]{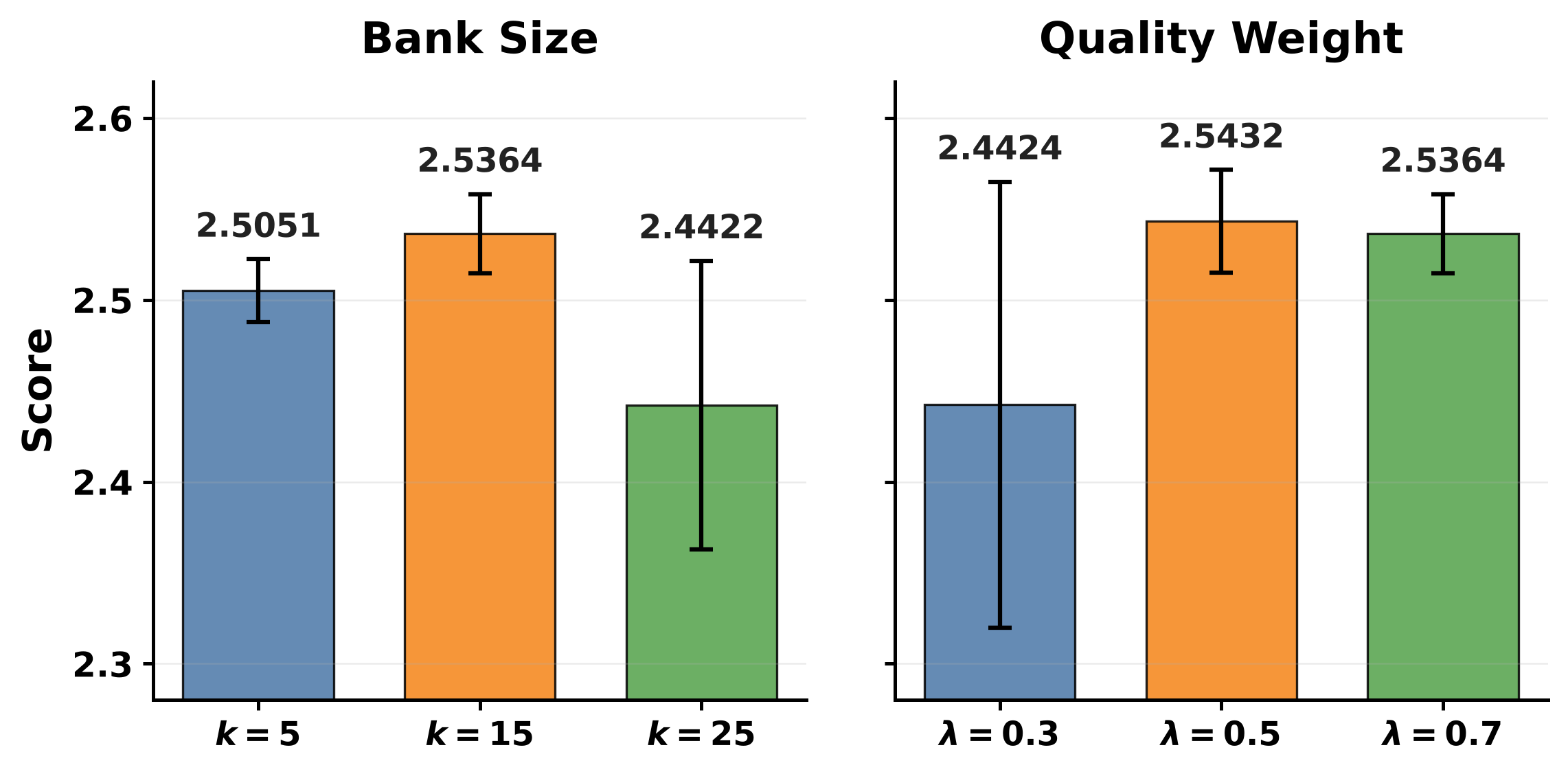}
    \caption{
Curation sensitivity on Circle Packing (Square).  Each setting reuses
the corresponding completed cheap pool and reruns curation and strong-model
refinement.  The default is $k=15$, $r=\text{min}(k, 10)$, and $\lambda=0.7$.
    }
\label{tab:curation_hparams}
\end{figure}



Figure \ref{tab:curation_hparams} reports the curation hyperparameter sensitivity on Circle Packing
(Square). Each setting reuses the same completed cheap-model candidate pool and
reruns only curation and strong-model refinement. The default configuration
($k=15$, $r=\text{min}(k, 10)$, $\lambda=0.7$) remains competitive across all tested settings.
For the bank/rank configurations, the intermediate setting performs best:
reducing the selected pool size limits coverage, while increasing it spreads
the strong-model refinement budget across more candidates. For the
quality-diversity trade-off, performance is stable around the default
$\lambda=0.7$, while emphasizing diversity more strongly ($\lambda=0.3$)
increases variance. We retain $\lambda=0.7$ and $k=15,r=10$ as the default
settings used in the main experiments.


\subsection{Cost Breakdown}

Table~\ref{tab:cost_breakdown} separates generation and representation costs.
The cheap-model column excludes embedding so that the three components are
disjoint. For every run, we retain the longest request-completion prefix that
does not exceed the task's 100\% dollar budget, matching the main table and
cost curves. Values are then averaged over the same three runs used in the
main table; percentages are computed relative to the mean total in the last
column.

\begin{table*}[t]
\centering
\caption{Mean \model cost by component. Parentheses give each
component's percentage of the recorded total cost.}
\label{tab:cost_breakdown}
\small
\begin{tabular}{lcccc}
\toprule
Task & Cheap generation & Strong generation & Embedding & Total \\
\midrule
Circle Packing (Square)
& \$0.1447 (12.10\%)
& \$1.0496 (87.73\%)
& \$0.00207 (0.17\%)
& \$1.1963 \\
Circle Packing (Rect)
& \$0.1463 (12.31\%)
& \$1.0419 (87.65\%)
& \$0.00054 (0.05\%)
& \$1.1887 \\
TXN Scheduling
& \$0.1423 (13.42\%)
& \$0.9172 (86.49\%)
& \$0.00098 (0.09\%)
& \$1.0605 \\
Prism
& \$0.1239 (11.33\%)
& \$0.9686 (88.56\%)
& \$0.00119 (0.11\%)
& \$1.0937 \\
\bottomrule
\end{tabular}
\end{table*}

Embedding accounts for only 0.05--0.17\% of recorded cost across tasks
(0.11\% after pooling the four task means). The embedding cache is shared
within a run, so a candidate's code and evaluator-behavior views are charged
only when they are first computed. The retained total can be slightly below
the nominal cutoff because requests are indivisible: we exclude the first
completed request that would cross the cutoff and all subsequent requests.
Raw terminal logs may be larger because requests already in flight are allowed
to finish, but those tails are excluded from both this table and all
budget-comparison scores.


\subsection{Handoff Lineage Case Study}

We trace the primary-parent lineage of the best final program in a Circle
Packing (Square) run. At handoff, the cheap phase contained 73 unique valid
candidates and \model selected 15 seeds. The final best program did
\emph{not} descend from the best cheap candidate. Its relay ancestor came
from cheap thread 3 (zero-indexed; the fourth instantiated trajectory), at
cheap generation 14. That candidate had normalized quality 0.6783 and ranked
ninth by quality both in the 73-candidate handoff pool and among the 15
selected seeds. By comparison, the best cheap candidate had quality 0.9022
and came from thread 2.

The rank-nine seed was inserted into the shared strong-model population and
was selected as the primary parent of the generation-21 program that achieved
the run's final best score: a sum of radii of 2.5364 (normalized score 0.9626).
Thus, a seed that appeared substantially weaker at handoff supported the best
strong-model refinement, whereas handing off only the early leader would have
discarded that lineage. This is mechanism-level evidence for population
handoff rather than single-leader continuation. It is consistent with the
quality--diversity curation ablation, but is not by itself a causal comparison
against quality-only top-$k$: the ancestor's rank also places it among the ten
mandatory high-quality anchors. The controlled curation results provide that
comparison.

\subsection{Different Model Family}

We repeat the Circle Packing (Square) comparison using GPT-OSS-20B as the cheap
model and GPT-OSS-120B as the strong model \cite{agarwal2025gpt}.  Input/output
prices are \$0.04/\$0.15 per million tokens for 20B and \$0.15/\$0.60 for
120B.  All experimental settings,
including budgets, baselines, and evaluation protocols, are kept identical to
the main Qwen-based experiments.

\begin{figure}[!t]
    \centering
\includegraphics[width=0.48\textwidth]{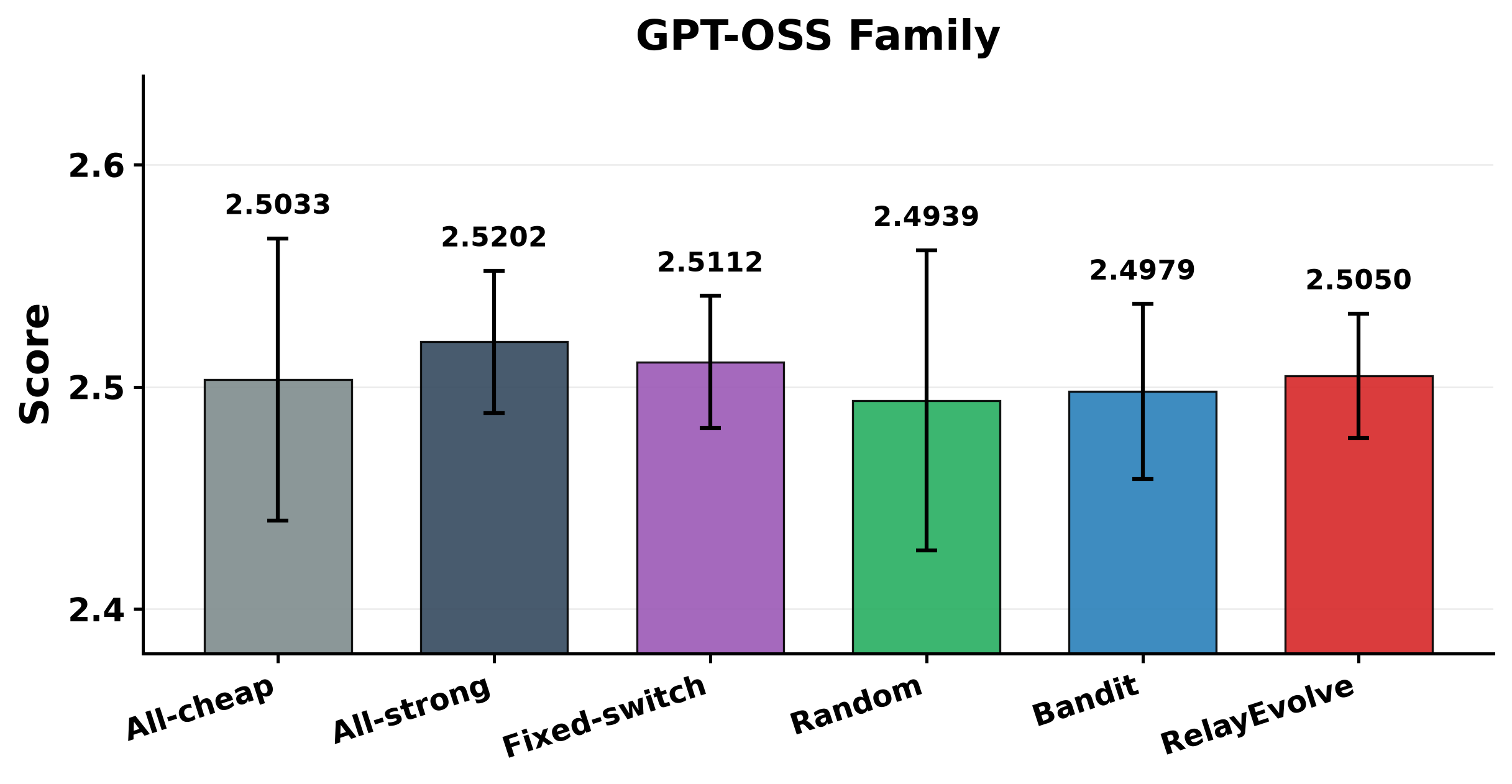}
    \caption{
Performance with the GPT-OSS model family  on
Circle Packing (Square). Bars show mean scores over three runs and error bars indicate sample
standard deviation.
    }
\label{tab:ada2}
\end{figure}

Figure \ref{tab:ada2} shows \model remains competitive and has lower variance than the other mixed
methods, but All-strong obtains the best mean for this model family.  This
negative result is informative: adaptive population handoff is not
uniformly superior under every cheap/strong pair, and its reward calibration
can depend on how much useful diversity the cheap model produces.  The main
Qwen results therefore establish the primary claim, while this experiment
provides evidence about its current boundary rather than an additional win.

\subsection{Different Evolutionary Backbone}

We evaluate whether \model can be transferred to a different
evolutionary backend by replacing the default ShinkaEvolve \cite{lange2025shinkaevolve} controller with
AdaEvolve \cite{cemri2026adaevolve}. We port the \model-specific components, including the
Grow--Deepen scheduler, relay bank, stopping rule, and curation stage, while
using AdaEvolve's population management, archive, parent selection, variation
operators, and evaluator pipeline. All methods use the same initial program,
task prompt, Qwen-3.5-Flash/Plus endpoints, and model prices. We use a fixed budget of \$0.5 for this comparison.
Due to the limited resource and additional  cost, we report two independent runs.

\begin{figure}[!t]
    \centering
\includegraphics[width=0.48\textwidth]{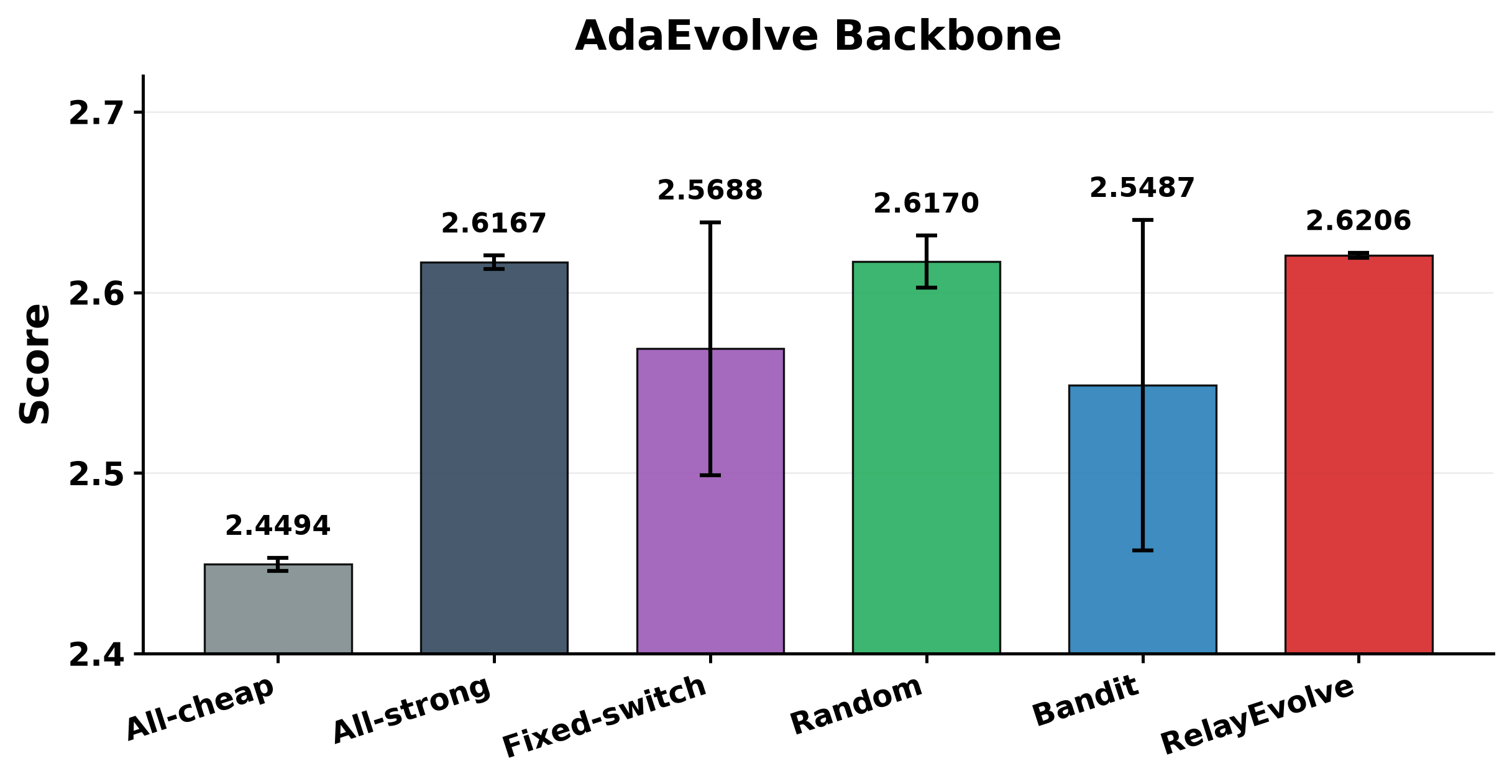}
    \caption{
Performance with the AdaEvolve evolutionary backbone on Circle
Packing (Square). Bars show the mean over two runs under \$0.5 budget, and error bars indicate
sample standard deviation.
    }
\label{tab:adaa}
\end{figure}

Figure~\ref{tab:adaa} shows that \model remains competitive after
switching evolutionary backends. Its two runs are also relatively consistent,
suggesting that the relay mechanism can operate with different population and
archive implementations. However, the result does not show universal
dominance: Random and All-strong remain competitive.

This result suggests that the benefit of
adaptive handoff depends on the interaction between the model pair and the
underlying evolutionary dynamics. The purpose of this experiment is therefore
to evaluate backbone compatibility rather than to establish another benchmark
win.

Note that LEVI is not included in this controlled comparison because its search procedure
is itself the baseline being evaluated. Replacing its internal evolutionary
procedure with AdaEvolve would no longer constitute the original LEVI method.
Its native-backbone results are reported separately.




\subsection{Discovered Program}

\begin{center}
\centering
\begin{tcolorbox}[
    colback=cyan!4,
    colframe=cyan!40!black,
    breakable,
    fonttitle=\bfseries\large,
    arc=1mm
]

\begin{lstlisting}[
    language=python,
    basicstyle=\ttfamily\scriptsize,
    keywordstyle=\color{blue}\bfseries,
    commentstyle=\color{green!50!black}\itshape,
    stringstyle=\color{orange!70!black},
    showstringspaces=false,
    breaklines=true,
    numbers=left,
    numberstyle=\tiny\color{gray},
    frame=none,
    xleftmargin=1em
]
# EVOLVE-BLOCK-START
"""Constructor-based circle packing for n=26 circles with priority constraint resolution"""
import numpy as np


class ConstraintManager:
    """
    Manages constraint propagation for circle packing with priority-based violation resolution.
    Processes circles in priority order based on border proximity and constraint tightness.
    """
    
    def __init__(self, centers):
        self.n = centers.shape[0]
        self.centers = centers
        self.border_dists = np.zeros(self.n)
        self.distances = np.zeros((self.n, self.n))
        self.neighbors = [[] for _ in range(self.n)]
        self._precompute_constraints()
    
    def _precompute_constraints(self):
        """Pre-compute border distances and pairwise distances."""
        for i in range(self.n):
            x, y = self.centers[i]
            self.border_dists[i] = min(x, y, 1 - x, 1 - y)
        
        for i in range(self.n):
            for j in range(i + 1, self.n):
                d = np.sqrt(np.sum((self.centers[i] - self.centers[j]) ** 2))
                self.distances[i, j] = self.distances[j, i] = d
                
                # Identify neighbors (circles that can interact)
                if d < 0.45:
                    self.neighbors[i].append(j)
                    self.neighbors[j].append(i)
    
    def compute_priorities(self, radii):
        """
        Calculate priority for each circle based on border proximity and constraint tightness.
        Priority = border_dist * 0.6 + (1 - constraint_score) * 0.4
        Lower priority value = higher processing priority
        """
        priorities = np.zeros(self.n)
        
        for i in range(self.n):
            # Border proximity component (lower border_dist = higher priority)
            border_component = self.border_dists[i]
            
            # Constraint tightness component
            constraint_score = 1.0
            if len(self.neighbors[i]) > 0:
                tightness_sum = 0.0
                for j in self.neighbors[i]:
                    if radii[i] + radii[j] > 0:
                        ratio = (radii[i] + radii[j]) / self.distances[i, j]
                        tightness_sum += min(ratio, 1.0)
                constraint_score = tightness_sum / len(self.neighbors[i])
            
            # Combined priority (lower = process first)
            priorities[i] = border_component * 0.6 + (1 - constraint_score) * 0.4
        
        return priorities
    
    def compute_max_radii(self, max_iterations=60, tolerance=1e-9):
        """
        Compute maximum valid radii using priority-based constraint propagation.
        """
        radii = self.border_dists.copy()
        
        for iteration in range(max_iterations):
            max_change = 0
            
            # Phase 1: Calculate priorities and identify violations
            priorities = self.compute_priorities(radii)
            violations = []
            
            for i in range(self.n):
                for j in self.neighbors[i]:
                    if j > i:  # Avoid duplicates
                        if radii[i] + radii[j] > self.distances[i, j] + 1e-12:
                            # Priority based on the more constrained circle
                            violation_priority = min(priorities[i], priorities[j])
                            violations.append((violation_priority, i, j))
            
            # Sort violations by priority (process border-adjacent first)
            violations.sort(key=lambda x: x[0])
            
            # Phase 2: Resolve violations in priority order
            for _, i, j in violations:
                dist = self.distances[i, j]
                if radii[i] + radii[j] > dist + 1e-12:
                    total = radii[i] + radii[j]
                    if total > 1e-12:
                        ratio_i = radii[i] / total
                        ratio_j = radii[j] / total
                        
                        new_radii_i = dist * ratio_i
                        new_radii_j = dist * ratio_j
                        
                        change = abs(new_radii_i - radii[i]) + abs(new_radii_j - radii[j])
                        max_change = max(max_change, change)
                        
                        radii[i] = new_radii_i
                        radii[j] = new_radii_j
            
            # Phase 3: Expand radii where possible (reverse priority order)
            expansion_order = np.argsort(-priorities)
            
            for i in expansion_order:
                max_r = self.border_dists[i]
                for j in self.neighbors[i]:
                    max_r = min(max_r, self.distances[i, j] - radii[j])
                
                if max_r > radii[i] + 1e-12:
                    change = max_r - radii[i]
                    max_change = max(max_change, change)
                    radii[i] = max_r
            
            if max_change < tolerance:
                break
        
        # Final constraint enforcement pass
        for i in range(self.n):
            radii[i] = min(radii[i], self.border_dists[i])
            for j in range(self.n):
                if i != j:
                    radii[i] = min(radii[i], self.distances[i, j] - radii[j])
        
        return radii


def construct_packing():
    """
    Construct a specific arrangement of 26 circles in a unit square
    that attempts to maximize the sum of their radii.

    Returns:
        Tuple of (centers, radii, sum_of_radii)
        centers: np.array of shape (26, 2) with (x, y) coordinates
        radii: np.array of shape (26) with radius of each circle
        sum_of_radii: Sum of all radii
    """
    n = 26
    centers = np.zeros((n, 2))
    
    # Optimized hexagonal packing with better spacing for n=26
    # 5-6-5-5-5 row configuration places wider row in middle
    row_spacing = 0.172
    col_spacing = 0.198
    offset = col_spacing * 0.5
    
    circle_idx = 0
    
    # Row 0: 5 circles along bottom
    for i in range(5):
        x = 0.092 + i * col_spacing
        y = 0.082
        centers[circle_idx] = [x, y]
        circle_idx += 1
    
    # Row 1: 6 circles, offset (wider row)
    for i in range(6):
        x = 0.072 + i * col_spacing * 0.93
        y = 0.082 + row_spacing
        centers[circle_idx] = [x, y]
        circle_idx += 1
    
    # Row 2: 5 circles, aligned with row 0
    for i in range(5):
        x = 0.092 + i * col_spacing
        y = 0.082 + 2 * row_spacing
        centers[circle_idx] = [x, y]
        circle_idx += 1
    
    # Row 3: 5 circles, offset
    for i in range(5):
        x = 0.092 + offset + i * col_spacing
        y = 0.082 + 3 * row_spacing
        centers[circle_idx] = [x, y]
        circle_idx += 1
    
    # Row 4: 5 circles along top
    for i in range(5):
        x = 0.092 + i * col_spacing
        y = 0.082 + 4 * row_spacing
        centers[circle_idx] = [x, y]
        circle_idx += 1
    
    # Compute maximum valid radii using ConstraintManager
    constraint_mgr = ConstraintManager(centers)
    radii = constraint_mgr.compute_max_radii()
    
    # Gradient-based position refinement with momentum
    centers, radii = refine_positions_momentum(centers, radii, max_iterations=100, learning_rate=0.01)
    
    sum_radii = np.sum(radii)
    
    return centers, radii, sum_radii


def refine_positions_momentum(centers, radii, max_iterations=100, learning_rate=0.01):
    """
    Refine circle positions using gradient-based optimization with momentum.
    Momentum helps smooth convergence and escape local minima.
    """
    n = centers.shape[0]
    centers = centers.copy()
    velocity = np.zeros_like(centers)
    
    initial_lr = learning_rate
    momentum = 0.7
    epsilon = 0.0015
    
    best_centers = centers.copy()
    best_sum = np.sum(radii)
    
    for iteration in range(max_iterations):
        # Adaptive learning rate with warm restarts
        if iteration % 30 == 0 and iteration > 0:
            initial_lr *= 0.8
        
        current_lr = initial_lr * (0.97 ** iteration)
        
        # Compute current objective
        constraint_mgr = ConstraintManager(centers)
        current_radii = constraint_mgr.compute_max_radii()
        current_sum = np.sum(current_radii)
        
        # Track best solution
        if current_sum > best_sum:
            best_sum = current_sum
            best_centers = centers.copy()
        
        # Compute gradient using finite differences
        gradients = np.zeros_like(centers)
        
        for i in range(n):
            for dim in range(2):
                # Central difference
                centers_plus = centers.copy()
                centers_plus[i, dim] = np.clip(centers_plus[i, dim] + epsilon, 0.04, 0.96)
                
                centers_minus = centers.copy()
                centers_minus[i, dim] = np.clip(centers_minus[i, dim] - epsilon, 0.04, 0.96)
                
                mgr_plus = ConstraintManager(centers_plus)
                mgr_minus = ConstraintManager(centers_minus)
                
                sum_plus = np.sum(mgr_plus.compute_max_radii())
                sum_minus = np.sum(mgr_minus.compute_max_radii())
                
                gradients[i, dim] = (sum_plus - sum_minus) / (2 * epsilon)
        
        # Apply momentum to velocity
        velocity = momentum * velocity + current_lr * gradients
        
        # Clip velocity to prevent instability
        vel_norm = np.sqrt(np.sum(velocity ** 2))
        if vel_norm > 0.3:
            velocity = velocity * 0.3 / vel_norm
        
        # Update positions
        centers += velocity
        centers = np.clip(centers, 0.05, 0.95)
        
        # Early termination
        if iteration > 30:
            new_radii = ConstraintManager(centers).compute_max_radii()
            new_sum = np.sum(new_radii)
            if abs(new_sum - current_sum) < 5e-7:
                break
    
    # Return best solution found
    final_mgr = ConstraintManager(best_centers)
    final_radii = final_mgr.compute_max_radii()
    
    return best_centers, final_radii


# EVOLVE-BLOCK-END


# This part remains fixed (not evolved)
def run_packing():
    """Run the circle packing constructor for n=26"""
    centers, radii, sum_radii = construct_packing()
    return centers, radii, sum_radii


def visualize(centers, radii):
    """
    Visualize the circle packing

    Args:
        centers: np.array of shape (n, 2) with (x, y) coordinates
        radii: np.array of shape (n) with radius of each circle
    """
    import matplotlib.pyplot as plt
    from matplotlib.patches import Circle

    fig, ax = plt.subplots(figsize=(8, 8))

    # Draw unit square
    ax.set_xlim(0, 1)
    ax.set_ylim(0, 1)
    ax.set_aspect("equal")
    ax.grid(True)

    # Draw circles
    for i, (center, radius) in enumerate(zip(centers, radii)):
        circle = Circle(center, radius, alpha=0.5)
        ax.add_patch(circle)
        ax.text(center[0], center[1], str(i), ha="center", va="center")

    plt.title(f"Circle Packing (n={len(centers)}, sum={sum(radii):.6f})")
    plt.show()


if __name__ == "__main__":
    centers, radii, sum_radii = run_packing()
    print(f"Sum of radii: {sum_radii}")
    # AlphaEvolve improved this to 2.635

    # Uncomment to visualize:
    visualize(centers, radii)
    
\end{lstlisting}

\end{tcolorbox}
\captionof{figure}{Best program discovered by \model  in Circle Packing (Square).}
\label{fig:best_prism}
\end{center}

\begin{center}
\centering
\begin{tcolorbox}[
    colback=cyan!4,
    colframe=cyan!40!black,
    breakable,
    fonttitle=\bfseries\large,
    arc=1mm
]

\begin{lstlisting}[
    language=python,
    basicstyle=\ttfamily\scriptsize,
    keywordstyle=\color{blue}\bfseries,
    commentstyle=\color{green!50!black}\itshape,
    stringstyle=\color{orange!70!black},
    showstringspaces=false,
    breaklines=true,
    numbers=left,
    numberstyle=\tiny\color{gray},
    frame=none,
    xleftmargin=1em
]
# EVOLVE-BLOCK-START
import numpy as np
from scipy.optimize import minimize


def circle_packing21() -> np.ndarray:
    """
    Places 21 non-overlapping circles inside a rectangle of perimeter 4.

    Uses hexagonal packing with adaptive aspect ratio search and local optimization
    to maximize sum of radii.
    """
    np.random.seed(42)

    perimeter_constraint = 2.0  # W + H = 2

    # Multiple row patterns to explore different packing arrangements
    row_patterns = [
        [4, 4, 5, 4, 4],      # 5 rows, symmetric
        [5, 4, 4, 4, 4],      # 5 rows
        [4, 5, 4, 4, 4],      # 5 rows
        [3, 4, 5, 5, 4],      # 5 rows, varied
        [4, 4, 4, 5, 4],      # 5 rows
        [5, 5, 4, 4, 3],      # 5 rows
        [3, 5, 5, 5, 3],      # 5 rows, symmetric
        [4, 5, 5, 4, 3],      # 5 rows
    ]

    best_config = None
    best_sum_radii = 0.0

    # Phase 1: Coarse aspect ratio search (10 ratios)
    coarse_aspects = np.linspace(0.9, 1.5, 10)

    # Phase 2: Fine search around top performers
    fine_aspects_per_peak = 13  # ~40 total fine ratios

    phase1_results = []

    # Phase 1: Coarse search
    for aspect in coarse_aspects:
        W = perimeter_constraint * aspect / (1 + aspect)
        H = perimeter_constraint - W

        if W <= 0 or H <= 0:
            continue

        for pattern in row_patterns:
            result = optimize_for_config(W, H, pattern)
            if result is not None:
                sum_r = np.sum(result[:, 2])
                phase1_results.append((sum_r, aspect, pattern, result))

    # Sort by sum_radii and get top 3 aspect ratios
    phase1_results.sort(key=lambda x: x[0], reverse=True)

    # Get unique top aspect ratios for fine search
    top_aspects = []
    for item in phase1_results[:10]:  # Check top 10 for diversity
        if item[1] not in top_aspects:
            top_aspects.append(item[1])
        if len(top_aspects) >= 3:
            break

    # Phase 2: Fine search around top aspect ratios
    fine_aspects = []
    for base_aspect in top_aspects:
        # Search in a narrow range around each top aspect
        fine_range = np.linspace(base_aspect - 0.03, base_aspect + 0.03, fine_aspects_per_peak)
        fine_aspects.extend(fine_range)

    # Remove duplicates and sort
    fine_aspects = sorted(list(set(fine_aspects)))

    # Run fine search
    for aspect in fine_aspects:
        W = perimeter_constraint * aspect / (1 + aspect)
        H = perimeter_constraint - W

        if W <= 0 or H <= 0:
            continue

        for pattern in row_patterns:
            result = optimize_for_config(W, H, pattern)
            if result is not None:
                sum_r = np.sum(result[:, 2])
                if sum_r > best_sum_radii:
                    best_sum_radii = sum_r
                    best_config = result.copy()

    if best_config is None:
        # Fallback configuration
        r = 0.1
        xs = r + np.arange(7) * 2 * r
        ys = r + np.arange(3) * 2 * r
        circles = []
        for y in ys:
            for x in xs:
                circles.append([float(x), float(y), float(r)])
        return np.array(circles, dtype=float)

    return best_config


def optimize_for_config(W, H, rows_config):
    """
    Optimize circle positions and radii for a given rectangle and row pattern.
    """
    # Initial radius estimate based on rectangle dimensions
    r_init = min(W, H) / 8.0

    # Generate hexagonal lattice positions
    circles = []
    y_offset = r_init
    for i, count in enumerate(rows_config):
        if count <= 0:
            continue
        if i % 2 == 0:
            # Even row: centered
            row_width = (count - 1) * 2 * r_init
            x_start = (W - row_width) / 2
            for j in range(count):
                circles.append([x_start + j * 2 * r_init, y_offset, r_init])
        else:
            # Odd row: offset
            row_width = (count - 1) * 2 * r_init
            x_start = (W - row_width) / 2 + r_init
            for j in range(count):
                circles.append([x_start + j * 2 * r_init, y_offset, r_init])
        y_offset += np.sqrt(3) * r_init

    if len(circles) != 21:
        return None

    # Convert to numpy array
    circles = np.array(circles, dtype=float)

    # Local optimization to maximize sum of radii
    def objective(params):
        radii = params[2::3]
        return -np.sum(radii)

    # Safety margin for constraint validation robustness
    margin = 1e-6

    def constraints(params):
        constraints_list = []
        for i in range(21):
            x, y, r = params[3*i], params[3*i+1], params[3*i+2]
            # Boundary constraints with safety margin
            constraints_list.append(x - r - margin)
            constraints_list.append(W - (x + r) - margin)
            constraints_list.append(y - r - margin)
            constraints_list.append(H - (y + r) - margin)

        for i in range(21):
            for j in range(i + 1, 21):
                xi, yi, ri = params[3*i], params[3*i+1], params[3*i+2]
                xj, yj, rj = params[3*j], params[3*j+1], params[3*j+2]
                dist = np.sqrt((xi - xj)**2 + (yi - yj)**2)
                # Non-overlap constraint with safety margin
                constraints_list.append(dist - (ri + rj) - margin)

        return np.array(constraints_list)

    bounds = []
    for i in range(21):
        bounds.append((0.001, W - 0.001))
        bounds.append((0.001, H - 0.001))
        bounds.append((0.001, min(W, H) / 2))

    x0 = circles.flatten()

    cons = {'type': 'ineq', 'fun': constraints}
    result = minimize(objective, x0, method='SLSQP', bounds=bounds,
                     constraints=cons, options={'maxiter': 1000, 'ftol': 1e-9})

    if not result.success:
        return None

    optimized_circles = np.zeros((21, 3), dtype=float)
    for i in range(21):
        optimized_circles[i] = [result.x[3*i], result.x[3*i+1], result.x[3*i+2]]

    # Verify all constraints are satisfied (with safety margin)
    margin = 1e-6
    for i in range(21):
        x, y, r = optimized_circles[i]
        if x < r - margin or x > W - r + margin or y < r - margin or y > H - r + margin:
            return None

    # Verify non-overlap constraints
    centers = optimized_circles[:, :2]
    radii = optimized_circles[:, 2]
    diff = centers[:, np.newaxis, :] - centers[np.newaxis, :, :]
    dists = np.sqrt(np.sum(diff**2, axis=2))
    min_sep = radii[:, np.newaxis] + radii[np.newaxis, :]
    mask = ~np.eye(21, dtype=bool)
    if np.any(dists[mask] < min_sep[mask] - margin):
        return None

    return optimized_circles


# EVOLVE-BLOCK-END

if __name__ == "__main__":
    circles = circle_packing21()
    print(f"Radii sum: {np.sum(circles[:, -1])}")

\end{lstlisting}

\end{tcolorbox}
\captionof{figure}{Best program discovered by \model in Circle Packing (Rect).}
\label{fig:best_prism}
\end{center}

\begin{center}
\centering
\begin{tcolorbox}[
    colback=cyan!4,
    colframe=cyan!40!black,
    breakable,
    fonttitle=\bfseries\large,
    arc=1mm
]

\begin{lstlisting}[
    language=python,
    basicstyle=\ttfamily\scriptsize,
    keywordstyle=\color{blue}\bfseries,
    commentstyle=\color{green!50!black}\itshape,
    stringstyle=\color{orange!70!black},
    showstringspaces=false,
    breaklines=true,
    numbers=left,
    numberstyle=\tiny\color{gray},
    frame=none,
    xleftmargin=1em
]
import random

from txn_simulator import Workload
from workloads import WORKLOAD_1, WORKLOAD_2, WORKLOAD_3

# EVOLVE-BLOCK-START

def get_best_schedule(workload, num_seqs):
    """
    Get optimal schedule using multi-start greedy with enhanced local search.

    Returns:
        Tuple of (lowest makespan, corresponding schedule)
    """
    best_overall_cost = float('inf')
    best_txn_seq = []

    def calculate_schedule_cost(seq):
        """Calculate makespan for a given transaction sequence."""
        return workload.get_opt_seq_cost(seq)

    def get_greedy_cost_sampled(num_samples):
        """Greedy construction with random sampling at each step."""
        start_txn = random.randint(0, workload.num_txns - 1)
        txn_seq = [start_txn]
        remaining_txns = [x for x in range(0, workload.num_txns)]
        remaining_txns.remove(start_txn)

        for i in range(0, workload.num_txns - 1):
            min_cost = float('inf')
            min_txn = -1
            holdout_txns = []

            # Sample candidates instead of evaluating all
            for j in range(0, num_samples):
                if len(remaining_txns) == 0:
                    break
                idx = random.randint(0, len(remaining_txns) - 1)
                t = remaining_txns.pop(idx)
                holdout_txns.append(t)

                test_seq = txn_seq.copy()
                test_seq.append(t)
                cost = calculate_schedule_cost(test_seq)

                if cost < min_cost:
                    min_cost = cost
                    min_txn = t

            # Restore remaining transactions
            for t in holdout_txns:
                remaining_txns.append(t)

            assert(min_txn != -1)
            txn_seq.append(min_txn)
            remaining_txns.remove(min_txn)

        overall_cost = calculate_schedule_cost(txn_seq)
        return overall_cost, txn_seq

    def conflict_aware_greedy():
        """Build schedule considering all available transactions."""
        seq = []
        available = list(range(workload.num_txns))

        while available:
            best_txn = None
            best_cost = float('inf')

            for txn in available:
                test_seq = seq + [txn]
                cost = calculate_schedule_cost(test_seq)

                if cost < best_cost:
                    best_cost = cost
                    best_txn = txn

            seq.append(best_txn)
            available.remove(best_txn)

        return seq, best_cost

    def local_search_enhanced(initial_seq, max_iterations=10):
        """Enhanced local search with multiple perturbation operators."""
        best_seq = initial_seq[:]
        best_cost = calculate_schedule_cost(best_seq)
        
        for iteration in range(max_iterations):
            improved = False
            
            # Try pairwise swaps (any two positions)
            for i in range(len(best_seq) - 1):
                for j in range(i + 1, len(best_seq)):
                    test_seq = best_seq[:]
                    test_seq[i], test_seq[j] = test_seq[j], test_seq[i]
                    cost = calculate_schedule_cost(test_seq)

                    if cost < best_cost:
                        best_cost = cost
                        best_seq = test_seq
                        improved = True
            
            # Try move operator (relocate transaction)
            for i in range(len(best_seq)):
                for pos in range(len(best_seq)):
                    if i != pos:
                        test_seq = best_seq[:]
                        moved = test_seq.pop(i)
                        test_seq.insert(pos, moved)
                        cost = calculate_schedule_cost(test_seq)
                        
                        if cost < best_cost:
                            best_cost = cost
                            best_seq = test_seq
                            improved = True
            
            if not improved:
                break
        
        return best_seq, best_cost

    def local_search_with_escape(initial_seq, max_iterations=50):
        """Local search with occasional acceptance of worse moves."""
        best_seq = initial_seq[:]
        best_cost = calculate_schedule_cost(best_seq)
        current_seq = best_seq[:]
        current_cost = best_cost
        
        for iteration in range(max_iterations):
            # Try three types of perturbations
            op = random.randint(0, 2)
            if op == 0:
                # Swap two random transactions
                i, j = random.sample(range(len(current_seq)), 2)
                new_seq = current_seq[:]
                new_seq[i], new_seq[j] = new_seq[j], new_seq[i]
            elif op == 1:
                # Move transaction to new position
                i = random.randint(0, len(current_seq) - 1)
                new_seq = current_seq[:]
                moved = new_seq.pop(i)
                pos = random.randint(0, len(new_seq))
                new_seq.insert(pos, moved)
            else:
                # Reverse a segment
                i, j = random.sample(range(len(current_seq)), 2)
                i, j = min(i, j), max(i, j)
                new_seq = current_seq[:]
                new_seq[i:j+1] = reversed(new_seq[i:j+1])
            
            # Validate and evaluate
            if len(set(new_seq)) == workload.num_txns:
                new_cost = calculate_schedule_cost(new_seq)
                if new_cost < best_cost:
                    best_cost = new_cost
                    best_seq = new_seq[:]
                    current_seq = new_seq[:]
                    current_cost = new_cost
                elif new_cost < current_cost or random.random() < 0.05:
                    # Accept if better than current or with small probability
                    if new_cost < current_cost * 1.05:
                        current_seq = new_seq[:]
                        current_cost = new_cost
        
        return best_seq, best_cost

    # Strategy 1: Conflict-aware greedy with enhanced local search
    seq1, cost1 = conflict_aware_greedy()
    seq1_refined, cost1_refined = local_search_enhanced(seq1, max_iterations=5)
    if cost1_refined < best_overall_cost:
        best_overall_cost = cost1_refined
        best_txn_seq = seq1_refined

    # Strategy 2: Multi-start sampled greedy with escape local search
    for _ in range(num_seqs):
        overall_cost, txn_seq = get_greedy_cost_sampled(50)
        
        # Apply enhanced local search with escape mechanism
        refined_seq, refined_cost = local_search_with_escape(txn_seq, max_iterations=30)
        
        if refined_cost < best_overall_cost:
            best_overall_cost = refined_cost
            best_txn_seq = refined_seq

    # Final refinement pass
    if len(best_txn_seq) > 0:
        final_seq, final_cost = local_search_enhanced(best_txn_seq, max_iterations=3)
        if final_cost < best_overall_cost:
            best_overall_cost = final_cost
            best_txn_seq = final_seq

    assert len(set(best_txn_seq)) == workload.num_txns

    return best_overall_cost, best_txn_seq

# EVOLVE-BLOCK-END

def get_random_costs():
    workload_size = 100
    workload = Workload(WORKLOAD_1)

    makespan1, schedule1 = get_best_schedule(workload, 10)
    cost1 = workload.get_opt_seq_cost(schedule1)

    workload2 = Workload(WORKLOAD_2)
    makespan2, schedule2 = get_best_schedule(workload2, 10)
    cost2 = workload2.get_opt_seq_cost(schedule2)

    workload3 = Workload(WORKLOAD_3)
    makespan3, schedule3 = get_best_schedule(workload3, 10)
    cost3 = workload3.get_opt_seq_cost(schedule3)
    print(cost1, cost2, cost3)
    return cost1 + cost2 + cost3, [schedule1, schedule2, schedule3]


if __name__ == "__main__":
    makespan, schedule = get_random_costs()
    print(f"Makespan: {makespan}")
\end{lstlisting}

\end{tcolorbox}
\captionof{figure}{Best program discovered by \model in TXN.}
\label{fig:best_prism}
\end{center}

\begin{center}
\centering
\begin{tcolorbox}[
    colback=cyan!4,
    colframe=cyan!40!black,
    breakable,
    fonttitle=\bfseries\large,
    arc=1mm
]

\begin{lstlisting}[
    language=python,
    basicstyle=\ttfamily\scriptsize,
    keywordstyle=\color{blue}\bfseries,
    commentstyle=\color{green!50!black}\itshape,
    stringstyle=\color{orange!70!black},
    showstringspaces=false,
    breaklines=true,
    numbers=left,
    numberstyle=\tiny\color{gray},
    frame=none,
    xleftmargin=1em
]
GPU_MEM_SIZE = 80 # GB

# EVOLVE-BLOCK-START

def compute_model_placement(gpu_num, models):
    """
    Compute a model placement that minimizes the maximum KVPR across all GPUs.
    Uses Iterative Binary Search with Local Opt Feedback for tighter bounds.

    Args:
        gpu_num: Number of GPUs
        models: List of models to place

    Returns:
        A placement of models to GPUs
    """

    def _check_feasibility(target_kvpr):
        """
        Check if models can be placed such that no GPU exceeds target_kvpr.
        Transforms constraint to: Load + target_kvpr * Size <= target_kvpr * MEM
        Uses Best Fit Decreasing for better packing quality
        """
        capacity = target_kvpr * GPU_MEM_SIZE
        # Sort by composite weight (Load + Pressure * Size) descending for better packing
        sorted_models = sorted(models, key=lambda m: (m.req_rate / m.slo) + target_kvpr * m.model_size, reverse=True)

        gpu_loads = [0.0 for _ in range(gpu_num)]
        gpu_sizes = [0.0 for _ in range(gpu_num)]
        placement = {gpu_id: [] for gpu_id in range(gpu_num)}

        for model in sorted_models:
            m_load = model.req_rate / model.slo
            m_size = model.model_size

            # Best Fit: Find GPU that minimizes slack while fitting
            best_idx = None
            min_slack = float('inf')

            for gpu_id in range(gpu_num):
                # Check transformed constraint: sum(Weight) <= Capacity
                current_weight = gpu_loads[gpu_id] + target_kvpr * gpu_sizes[gpu_id]
                model_weight = m_load + target_kvpr * m_size

                if current_weight + model_weight <= capacity:
                    slack = capacity - (current_weight + model_weight)
                    if slack < min_slack:
                        min_slack = slack
                        best_idx = gpu_id

            if best_idx is not None:
                placement[best_idx].append(model)
                gpu_loads[best_idx] += m_load
                gpu_sizes[best_idx] += m_size
            else:
                # Not found any valid GPU, return failure
                return False, None

        return True, placement

    def _calculate_actual_kvpr(placement_dict):
        """Calculate the actual max KVPR of a placement."""
        max_kvpr = 0.0
        for gpu_id in range(gpu_num):
            if len(placement_dict[gpu_id]) == 0:
                continue
            total_req_rate = sum(m.req_rate / m.slo for m in placement_dict[gpu_id])
            total_model_size = sum(m.model_size for m in placement_dict[gpu_id])
            remaining_mem = GPU_MEM_SIZE - total_model_size
            if remaining_mem > 0:
                kvpr = total_req_rate / remaining_mem
                max_kvpr = max(max_kvpr, kvpr)
            else:
                return float('inf')
        return max_kvpr

    # Calculate bounds for binary search
    total_load = sum(m.req_rate / m.slo for m in models)
    total_size = sum(m.model_size for m in models)

    # Tight lower bound: average load/size ratio across all GPUs
    low = total_load / (gpu_num * GPU_MEM_SIZE) if (gpu_num * GPU_MEM_SIZE) > 0 else 0.0
    high = min(1000.0, max(total_load / GPU_MEM_SIZE * 2, 1.0))
    best_placement = None
    best_kvpr = float('inf')

    # Iterative binary search with local opt feedback (3 iterations)
    for iteration in range(3):
        # Binary search for minimum feasible max_KVPR
        search_low = low
        search_high = high
        iteration_best = None

        for _ in range(25):
            mid = (search_low + search_high) / 2
            feasible, placement = _check_feasibility(mid)

            if feasible:
                # Apply local optimization to get actual KVPR
                optimized_placement = _local_optimize_placement(placement, models, gpu_num)
                actual_kvpr = _calculate_actual_kvpr(optimized_placement)

                if actual_kvpr < best_kvpr:
                    best_kvpr = actual_kvpr
                    best_placement = optimized_placement

                # Use actual KVPR to tighten upper bound (feedback loop)
                search_high = mid
                iteration_best = optimized_placement
            else:
                search_low = mid

        # Update global bounds based on achieved KVPR (feedback)
        if best_placement is not None:
            low = max(low, best_kvpr * 0.9)  # Tighten lower bound
            high = min(high, best_kvpr * 1.1)  # Tighten upper bound

    if best_placement is None:
        # Fallback: try one more time with relaxed bounds
        feasible, best_placement = _check_feasibility(high * 1.5)
        if feasible:
            best_placement = _local_optimize_placement(best_placement, models, gpu_num)
        else:
            raise ValueError("Unable to place all models within memory constraints.")

    return best_placement

def _local_optimize_placement(placement, models, gpu_num):
    """
    Try to improve placement by swapping or moving models between GPUs to minimize max KVPR.
    """
    improved = True
    max_iterations = 100
    iteration = 0

    while improved and iteration < max_iterations:
        improved = False
        iteration += 1

        # Calculate current KVPR for each GPU
        def compute_kvpr(placement_dict):
            kvpr_list = []
            for gpu_id in range(gpu_num):
                if len(placement_dict[gpu_id]) == 0:
                    kvpr_list.append(0.0)
                else:
                    total_req_rate = sum(m.req_rate / m.slo for m in placement_dict[gpu_id])
                    total_model_size = sum(m.model_size for m in placement_dict[gpu_id])
                    remaining_mem = GPU_MEM_SIZE - total_model_size
                    if remaining_mem > 0:
                        kvpr_list.append(total_req_rate / remaining_mem)
                    else:
                        kvpr_list.append(float('inf'))
            return kvpr_list

        def is_valid_placement(placement_dict):
            for g in range(gpu_num):
                total_size = sum(m.model_size for m in placement_dict[g])
                if total_size > GPU_MEM_SIZE:
                    return False
            return True

        kvpr_per_gpu = compute_kvpr(placement)
        max_kvpr = max(kvpr_per_gpu)

        # Try swapping each model with models on other GPUs
        for gpu_id in range(gpu_num):
            for i, model in enumerate(placement[gpu_id]):
                for other_gpu_id in range(gpu_num):
                    if other_gpu_id == gpu_id:
                        continue

                    # Try moving model to other GPU (without swap)
                    new_placement = {g: list(pl) for g, pl in placement.items()}
                    new_placement[gpu_id].pop(i)
                    new_placement[other_gpu_id].append(model)

                    if is_valid_placement(new_placement):
                        new_kvpr_per_gpu = compute_kvpr(new_placement)
                        new_max_kvpr = max(new_kvpr_per_gpu)
                        if new_max_kvpr < max_kvpr:
                            placement = new_placement
                            max_kvpr = new_max_kvpr
                            improved = True
                            break

                    # Try swapping model with models on other GPU
                    for j, other_model in enumerate(placement[other_gpu_id]):
                        new_placement = {g: list(pl) for g, pl in placement.items()}
                        new_placement[gpu_id].pop(i)
                        new_placement[other_gpu_id].pop(j)
                        new_placement[gpu_id].append(other_model)
                        new_placement[other_gpu_id].append(model)

                        if is_valid_placement(new_placement):
                            new_kvpr_per_gpu = compute_kvpr(new_placement)
                            new_max_kvpr = max(new_kvpr_per_gpu)
                            if new_max_kvpr < max_kvpr:
                                placement = new_placement
                                max_kvpr = new_max_kvpr
                                improved = True
                                break

                    if improved:
                        break

                if improved:
                    break

            if improved:
                break

    return placement

# EVOLVE-BLOCK-END


if __name__ == "__main__":
    # Test the algorithm

    from evaluator import generate_test_gpu_models
    from evaluator import calculate_kvcache_pressure
    from evaluator import safe_float
    import numpy as np

    test_cases = generate_test_gpu_models()
    all_kvpr = []
    for i, (gpu_num, gpu_models) in enumerate(test_cases):

        results = compute_model_placement(gpu_num, gpu_models)
        max_kvpr = calculate_kvcache_pressure(results)
        all_kvpr.append(safe_float(max_kvpr))

    avg_kvpr = np.mean(all_kvpr)
    if avg_kvpr != 0:
        avg_kvpr = 1.0 / avg_kvpr


    print(f"Max KVPR: {avg_kvpr:.3f}")
\end{lstlisting}

\end{tcolorbox}
\captionof{figure}{Best program discovered by \model in Prism.}
\label{fig:best_prism}
\end{center}

\bibliography{aaai2027}